\documentclass[letterpaper,runningheads]{llncs}

\usepackage[mobile]{eccv}
\usepackage{eccvabbrv}

\usepackage{graphicx}
\usepackage{booktabs}
\usepackage{etoolbox}
\usepackage[tracking=true]{microtype}

\def\PSNRColName{\textls[-20]{PSNR\kern-0.05em$\uparrow$}}
\def\SSIMColName{\textls[-20]{SSIM\kern-0.05em$\uparrow$}}
\def\LPIPSColName{\textls[-40]{LPIPS\kern-0.05em$\downarrow$}}

\makeatletter
\newcommand{\ifempty}[3]{%
  \ifx#1\@empty #2\else #3\fi
}
\makeatother

\makeatletter
\newcommand{\iflabel}[4]{%
  \ifx#1\@empty
    \ifnum\expandafter#2=0
      #3
    \else
      #3
    \fi
  \else
    \ifnum\expandafter#2=1
      #4
    \else
      #3
    \fi
  \fi
}
\makeatother

\usepackage{hyperref}
\hypersetup{
  pdfborder = {0 0 0}, %
}

\def\codepublicationnote{The source code of our implementation is available at \url{https://github.com/SpectacularAI/3dgs-deblur}.
Additionally, the smartphone data set can be accessed at \url{https://doi.org/10.5281/zenodo.10848124} and our Deblur-NeRF data set variant at
\url{https://doi.org/10.5281/zenodo.10847884}.}

\newcommand{\orcidlink}[1]{}

\usepackage[square,numbers,sort&compress]{natbib}
\usepackage{multirow}

\usepackage[dvipsnames]{xcolor}

\renewcommand{\paragraph}[1]{\medskip\noindent\textbf{#1}~~}

\usepackage{tikz}
\usepackage{pgfplots}
\usetikzlibrary{hobby, fit, shapes.geometric, patterns, calc, math, arrows}
\usetikzlibrary{3d, shapes.arrows}
\pgfplotsset{compat=newest}

\newlength{\figurewidth}
\newlength{\figureheight}
\newlength{\figurerowheight}

\usepackage{pifont}

\usepackage{changepage}

\pgfplotsset{axis on top, scale only axis, y grid style={line width=.1pt, draw=gray!10,dashed},x grid style={line width=.1pt, draw=gray!10,dashed}}

\DeclareRobustCommand{\num}[1]{{\bf\scriptsize #1}.}

\title{Gaussian Splatting on the Move: \\ Blur and Rolling Shutter Compensation for Natural Camera Motion}
\author{Otto Seiskari$^1$\orcidlink{0000-0001-6207-5671} \and Jerry Ylilammi$^1$ \and Valtteri Kaatrasalo$^1$ \and Pekka Rantalankila$^1$ \and Matias Turkulainen$^{2,3}$\orcidlink{0009-0007-6931-2386} \and Juho Kannala$^{1,3,4}$\orcidlink{0000-0001-5088-4041} \and Esa Rahtu$^5$\orcidlink{0000-0001-8767-0864} \and Arno Solin$^{1,3}$\orcidlink{0000-0002-0958-7886}}
\authorrunning{Seiskari et al.}
\titlerunning{Gaussian Splatting on the Move}
\institute{$^1$~Spectacular AI, $^2$~ETH Zurich, $^3$~Aalto University\\$^4$~University of Oulu, $^5$~Tampere University\\ Corresponding author: {\footnotesize\tt otto.seiskari@spectacularai.com}}

\begin{document}

\maketitle
\begin{abstract}%
High-quality scene reconstruction and novel view synthesis based on Gaussian Splatting (3DGS) typically require steady, high-quality photographs, often impractical to capture with handheld cameras. We present a method that adapts to camera motion and allows high-quality scene reconstruction with handheld video data suffering from motion blur and rolling shutter distortion. Our approach is based on detailed modelling of the physical image formation process and utilizes velocities estimated using visual-inertial odometry (VIO). Camera poses are considered non-static during the exposure time of a single image frame and camera poses are further optimized in the reconstruction process. We formulate a differentiable rendering pipeline that leverages screen space approximation to efficiently incorporate rolling-shutter and motion blur effects into the 3DGS framework. Our results with both synthetic and real data demonstrate superior performance in mitigating camera motion over existing methods, thereby advancing 3DGS in naturalistic settings.\looseness-1
\end{abstract}
\noindent
\begin{minipage}{1.0\textwidth}
  \centering
  \begin{tikzpicture}[inner sep=0,outer sep=0]

  \foreach \file/\cap [count=\i] in {ficus/Input,ficus-mb/Motion blur,ficus-rs/Rolling shutter,ficus/Our clean render} {

    \fill[fill=black!05,draw=none] (.2*\i*\textwidth-.01\textwidth,-.06\textwidth) ellipse (.6 and .4);

    \node[minimum width=.2\textwidth,minimum height=.25\textwidth] (p\i) at (.2*\i*\textwidth,0) {\includegraphics[height=.2\textwidth,trim=0 10 0 10,clip]{fig/\file-alpha}};

    \node[font=\scriptsize,align=center] at (.2*\i*\textwidth-.01\textwidth,-.12\textwidth) {\vphantom{p}\cap};

  }

  \node[rotate=110,font=\tiny\it,scale=.8] at ($(p3) - (.7cm,0)$) {\color{gray}Rolling shutter effects};
  \node[rotate=90,font=\tiny,scale=.8] at ($(p2) - (.7cm,-.01)$) {\color{gray}Motion blur};
  \node[rotate=90,font=\tiny,scale=.8] at ($(p2) - (.703cm,.002)$) {\color{gray!50}Motion blur};
  \node[rotate=90,font=\tiny,scale=.8] at ($(p2) - (.701cm,.02)$) {\color{gray!50}Motion blur};

  \node at ($(p1)$) {\includegraphics[height=.25\textwidth,trim=0 10 0 10,clip]{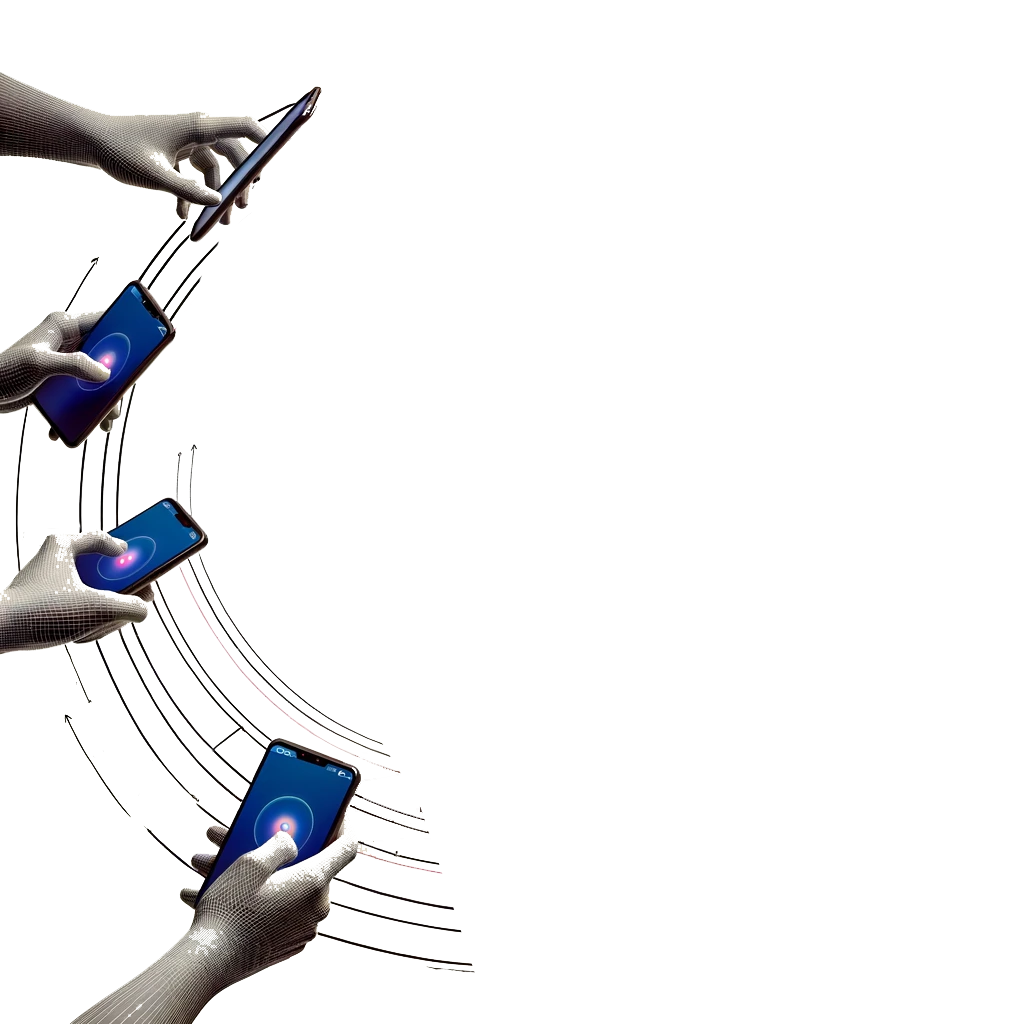}};

  \draw[black] ($(p1) + (-1.5cm,-1.5cm)$) edge[->,bend left=30] node[rotate=105,font=\tiny,fill=white,inner sep=2pt]{VIO} ++(-.5,2);

  \end{tikzpicture}
\end{minipage}
\section{Introduction}
\label{sec:introduction}
The field of novel view synthesis has seen significant advancements in recent years, with the introduction of Neural Radiance Fields (NeRF, \cite{NeRF}) and more recently, Gaussian Splatting (3DGS, \cite{GaussianSplatting}). Both classes of methods represent scenes as differentiable, non-mesh-based, 3D representations that allow rendering of new views that are often visually indistinguishable from evaluation images. One major limitation of these methods is that they generally require high-quality still photographs which can be accurately registered using photogrammetry software, such as COLMAP~\cite{COLMAP-1}. \newline
\indent Generating 3D reconstructions from casually captured data recorded with a moving hand-held camera, such as a smartphone, would enable faster data collection and integration into a broader range of use cases. However, image data from a moving sensor is prone to motion blur and rolling shutter distortion, which significantly degrade the quality of the reconstruction and increases the likelihood of failures in pose registration. Motion blur effects occur during the camera shutter's opening time due to relative motion between the camera and objects in the scene. Similarly, rolling shutter effects, caused by the camera sensor scanning the scene line-by-line, lead to warping distortions in fast-moving scenes or during rapid camera maneuvers. \newline
\indent Most of the existing methods aiming to compensate for these imaging effects use classical or deep learning--based approaches to recover sharpened versions of the input images, without the aid of an underlying 3D image formation model for the scene. The recent 3D novel view synthesis methods, including NeRF ~\cite{NeRF} and 3DGS ~\cite{GaussianSplatting}, allow for an alternative approach where a sharp 3D reconstruction is recovered without manipulating the training image data as an intermediary step. Deblur-NeRF~\cite{Deblur-NeRF}, and BAD-NeRF~\cite{wang2023badnerf} perform this in the context of NeRFs. In the context of 3DGS, \cite{lee2024deblurring} propose a method for a related but different problem of \emph{de-focus} blur compensation. Similarly to~\cite{wang2023badnerf}, our work focuses on \emph{camera motion blur} and applications to real data where this is the most prominent blurring modality, such as hand-held captures of static scenes without extreme close-up shots. \newline
\indent This work offers an alternative deblurring and rolling shutter correction approach for the 3DGS framework. Instead of learning blurring kernels from the data, as in \cite{Deblur-NeRF}, we directly model the image formation process with camera motion and rolling shutter effects leveraging velocity estimates computed using visual-inertial odometry (VIO), a technique that fuses inertial measurement unit (IMU) data with monocular video. We formulate an efficient differentiable motion blur and rolling-shutter capable rendering pipeline that utilizes screen space approximation to avoid recomputing or hindering the performance of 3DGS operations. To address the ill-posed nature of the deblurring problem (\cf~\cite{kaipio-somersalo-inverse-2004}), we leverage the regularization capabilities of the differentiable 3DGS framework priors information from sensor data. \newline
\indent The performance of our method is evaluated using both synthetic data from the Deblur-NeRF data set \cite{Deblur-NeRF} as well as real-world data recorded using mobile devices.
Our method is implemented as an extension to the NerfStudio \cite{Nerfstudio} and \texttt{gsplat} \cite{gsplat} software packages, which serve as our baseline methods for evaluation. Our approach consistently outperforms the baselines for both synthetic and real data experiments in terms of PSNR, SSIM, and LPIPS metrics, and the resulting reconstructions appear qualitatively sharper. \newline

\begin{figure}[t!]
  \centering\scriptsize
  \begin{tikzpicture}[inner sep=0]

    \setlength{\figurewidth}{.32\textwidth}
    \setlength{\figureheight}{.75\figurewidth}
    \setlength{\figurerowheight}{-1.02\figureheight}

    \node[anchor=center, rotate=90] (label-top) at (-1em,0.5\figurerowheight) {\bf Baseline};
    \node[anchor=center, rotate=90] (label-top) at (-1em,1.5\figurerowheight) {\bf Ours};

    \foreach \file/\label/\row/\col [count=\i] in {
        synthetic-mb-baseline/Motion blur/0/0,
        synthetic-rs-baseline/Rolling shutter/0/1,
        synthetic-posenoise-baseline/Pose noise/0/2,
        synthetic-mb-motion_blur/Motion blur/1/0,
        synthetic-rs-rolling_shutter/Rolling shutter/1/1,
        synthetic-posenoise-pose_opt/Pose noise/1/2
      } {
        \begin{scope}

        \node[anchor=north west, inner sep=2pt] (label-\i) at (\col*\figurewidth,\row*\figurerowheight) {\tiny\vphantom{g}\label};
        \clip[rounded corners=2pt] (label-\i.south west) -- (label-\i.south east) -- (label-\i.north east) -- (\col*\figurewidth+.99\figurewidth,\figurerowheight*\row) -- (\col*\figurewidth+.99\figurewidth,\figurerowheight*\row-\figureheight) -- (\col*\figurewidth,\figurerowheight*\row-\figureheight) -- cycle;

        \node [inner sep=0,minimum width=\figurewidth,minimum height=\figureheight,fill=red!10!white,anchor=north west] at (\col*\figurewidth, \row*\figurerowheight) {%
            \includegraphics[height=\figureheight]{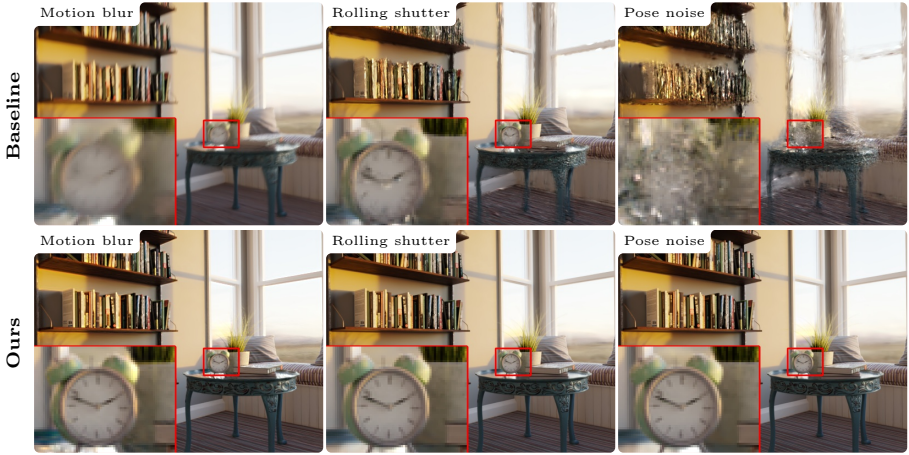}%
        };

        \end{scope}
    }

  \end{tikzpicture}
  \caption{3DGS reconstructions from the synthetic \texttt{cozyroom} scene under different simulated effects with (bottom row) and without (top row) our compensation. The corresponding numerical results are given in \cref{tab:synthetic}. The bottom right image is visually indistinguishable from the reference (PSNR 36.2).}
  \label{fig:synthetic}
\end{figure}

\begin{figure}[t]
  \centering\scriptsize

  \begin{tikzpicture}[inner sep=0]

    \setlength{\figurewidth}{.32\textwidth}
    \setlength{\figureheight}{.75\figurewidth}
    \setlength{\figurerowheight}{-1.02\figureheight}

    \node[anchor=center, rotate=90] (label-top) at (-1em,0.5\figurerowheight) {\bf Baseline};
    \node[anchor=center, rotate=90] (label-top) at (-1em,1.5\figurerowheight) {\bf Ours};

    \foreach \file/\label/\row/\col [count=\i] in {
        synthetic-factory-mb-baseline/Motion blur/0/0,
        synthetic-factory-rs-baseline/Rolling shutter/0/1,
        synthetic-factory-posenoise-baseline/Pose noise/0/2,
        synthetic-factory-mb-motion_blur/Motion blur/1/0,
        synthetic-factory-rs-rolling_shutter/Rolling shutter/1/1,
        synthetic-factory-posenoise-pose_opt/Pose noise/1/2
      } {
        \begin{scope}

        \node[anchor=north west, inner sep=2pt] (label-\i) at (\col*\figurewidth,\row*\figurerowheight) {\tiny\vphantom{g}\label};
        \clip[rounded corners=2pt] (label-\i.south west) -- (label-\i.south east) -- (label-\i.north east) -- (\col*\figurewidth+.99\figurewidth,\figurerowheight*\row) -- (\col*\figurewidth+.99\figurewidth,\figurerowheight*\row-\figureheight) -- (\col*\figurewidth,\figurerowheight*\row-\figureheight) -- cycle;

        \node [inner sep=0,minimum width=\figurewidth,minimum height=\figureheight,fill=red!10!white,anchor=north west] at (\col*\figurewidth, \row*\figurerowheight) {%
            \includegraphics[height=\figureheight]{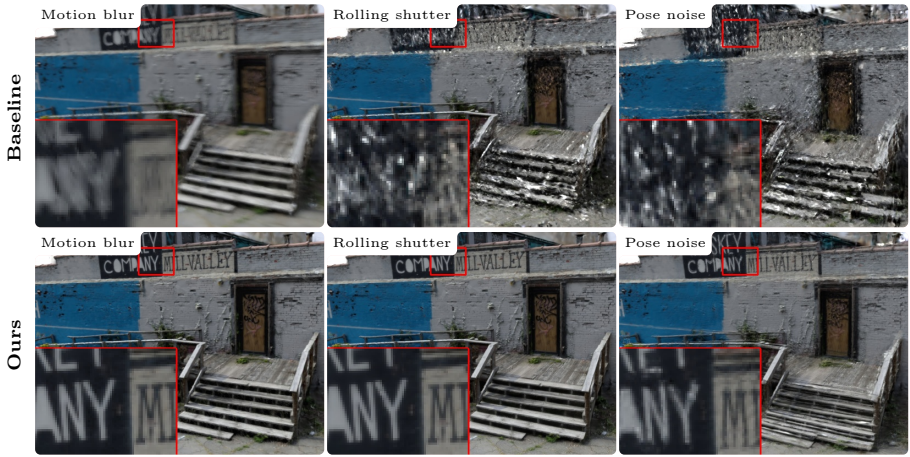}%
        };

        \end{scope}
    }

  \end{tikzpicture}
  \caption{Further 3DGS reconstructions from the synthetic \texttt{factory} scene. The Splatfacto method acting as a baseline. The corresponding numerical results are given in \cref{tab:synthetic}. In this case, the pose optimization case converged to a local minimum.}
  \label{fig:synthetic-factory}
\end{figure}

\section{Related Work}
\label{sec:related}
We cover classical and deep learning--based image deblurring and give an overview of prior methods compensating for motion blur and rolling shutter distortion in the context of 3D reconstruction using implicit representations.

\paragraph{Image deblurring}%
Motion blur compensation has been extensively studied in single and multi-image settings. Classical approaches attempt to jointly recover sharp input images along with blurring kernels with optimization based methods \cite{blind-motion-deblurring-2007}. This task is ill-posed since multiple blurring kernels can result in the same blurred image, and prior work add regularization to account for this \cite{xu2010two, shan2008high, cai2009blind}. Richardson--Lucy deblurring \cite{projective-deblur} attempts to account for spatially varying blurring with projective homographies to describe three-dimensional camera motion. Deep learning-based methods (\eg, \cite{neural-motion-deblurring-2016, kupyn2018deblurgan, gong2017blur2mf}) utilize features learned on large training data sets to recover sharp images. These methods outperform classical approaches that rely on hand crafted image statistics and can better handle spatially varying deblurring.

Similarly, rolling shutter (RS) compensation has been traditionally studied as a problem of optimally warping individual pixels or images to account for the row-wise exposure of frames \cite{liang2008rs,calibration-free-rolling-shutter-2012,bowstoarrows-rolling-shutter-2016,robust-strong-rolling-shutter-2018}. Modern methods such as \cite{occlusion-aware-rs-2018, DSUN-2020, CVR-2022} utilize information from three-dimensional image formation and camera motion to better compensate for RS effects. Moreover, deep-learning based methods trained on large data sets demonstrate occlusion and in-painting capability, a feature that traditional methods lack. Rolling-shutter compensation has also been studied in the scope of Structure-from-Motion (SfM) and Visual-Inertial SLAM methods \cite{rolling-shutter-ba-2012,spline-fusion-2013,rolling-shutter-vio-2019,rolling-shutter-ba-2022}.

Simultaneous motion blur and rolling shutter compensation has also been studied, in, \eg, \cite{rs-mb-2015, rs-mb-2017}. Learning-based methods aimed at generic image blur and artefacts, such as \cite{multi-stage-restoration-2021}, have also improved significantly and were used in \cite{Deblur-NeRF} as a pre-processing step to enhance NeRF reconstruction.

\paragraph{Deblurring 3D implicit representations}
Deblurring differentiable implicit representations for novel view synthesis is a relatively new topic and has been previously studied in \cite{Deblur-NeRF, hybrid-neural-motion-blur-2023, wang2023badnerf} for NeRF \cite{NeRF} representations and \cite{lee2024deblurring} in the 3DGS \cite{GaussianSplatting} context. Both \cite{Deblur-NeRF} and \cite{lee2024deblurring} incorporate additional learnable parameters that model blurring effects as part of the rendering pipeline in the form of small Multi-Layer Perceptrons (MLPs) added to the baseline models. These methods treat the training images as fixed in time and focus on extracting sharper images from blurred inputs. An alternative method is to model the per frame image formation model by accounting for the motion trajectory during the capture process as presented in \cite{wang2023badnerf}. Our approach is most similar to \cite{wang2023badnerf} in the sense that we explicitly model the blur formation process by integrating information over a short camera trajectory. Rolling shutter effects, which are not considered in the aforementioned works, is separately considered in the NeRF context in \cite{li2024usbnerf}, which also incorporates additional learnable parameters to the training process. In contrast, our approach does not include additional MLPs to the 3DGS pipeline, but directly models the local camera trajectory using linear and angular velocities, for which good initial estimates are readily available from IMU data captured from Visual-Inertial Odometry (VIO) pipelines.

Our method also utilizes pose optimization in the 3DGS framework, primarily for better pose registration in the presence of rolling shutter effects, which are not efficiently handled by COLMAP~\cite{COLMAP-1}. Pose refinement can also contribute to sharpening the reconstruction quality by effectively mitigating deblurring caused by the sensitivity of NeRF and 3DGS based methods on accurate pose estimates. Pose refinement has been previously applied to 3DGS in \cite{GS-SLAM, SplatTAM, matsuki2023gaussian, fu2023colmapfree}, but not together with rolling shutter compensation. Correspondingly, pose optimization with NeRFs has been studied in several works (\eg,~\cite{park2023camp, wang2023badnerf}), starting with BARF~\cite{BARF}. Utilizing information from VIO in NeRFs to stabilize pose optimization has been studied in \cite{vio-barf-2022}, but to the best of our knowledge, has not been studied in the 3DGS context.

Our implementation is based on Nerfstudio \cite{Nerfstudio} and \texttt{gsplat} \cite{gsplat}. For robust VIO, we utilize the Spectacular AI SDK, a proprietary VISLAM system loosely based on~\cite{HybVIO}.

\section{Methods}
\label{sec:methods}
We start with a brief overview of Gaussian Splatting and establishing notation, followed by our formulation of motion blur and rolling shutter effects in the 3DGS framework. For this, we also introduce a novel screen space approximation and pixel velocity-based rasterization, concluding with pose optimization and its role in the evaluation methodology.

\subsection{Gaussian Splatting}
Gaussian Splatting (3DGS, \cite{GaussianSplatting}) serves as an example and current cornerstone of \emph{differentiable rendering}, enabling the differentiation of a pixel's colour in relation to model parameters. This capability allows for the optimization of a loss function by comparing rendered images to real reference images. In essence, Gaussian Splatting maps the 3D scene into 2D images through a set of Gaussian distributions with additional colour and transparency attributes, each contributing to the region of the scene defined by the its mean and covariance.

In Gaussian Splatting, the colour
\begin{equation}
\label{eq:original-rendering-model}
  C_i(x, y, P_i, \mathcal G) \in \mathcal C
\end{equation}
of the pixel $(x, y) \in [0, W) \times [0, H)$ of output image $i$ can be differentiated with respect to the model parameters $\mathcal G$, and these parameters can be used to optimize a loss function
\begin{equation}
  \mathcal G \mapsto \sum_{i=1}^{N_{\rm img}} \mathcal L\left[C_i(\cdot, \cdot, P_i, \mathcal G), C_i' \right]
\end{equation}
comparing the rendered images $C_i$ to a set of $N_{\rm img}$ real reference images $C_i'$. We denote by $P_i \in \mathrm{SE}(3)$ camera pose corresponding to image $i$ and assume $\mathcal C = \mathbb R^3$ for RGB colours.

The 3DGS model parameters consist of a set
\begin{equation*}
  \mathcal G = \{ (\mu_j, \Sigma_j, \alpha_j, \theta_j ) \}_j
\end{equation*}
of Gaussian distributions $(\mu_j, \Sigma_j)$, transparencies $\alpha_j$ and view-dependent colours $\theta_j \in \mathcal C^{N_{\rm sh}}$ represented by $N_{\rm sh}$ spherical harmonic coefficient vectors.

\subsection{Blur and Rolling Shutter as Camera Motion}
We simultaneously model motion blur and rolling shutter as dynamic three-dimensional effects caused by the motion of the camera along a continuous trajectory $t \mapsto P(t)$. In terms of 3DGS, this can be modelled by changing the rendering equation to
\begin{equation}
\label{eq:motion-blur-and-rolling-shutter}
  C_i(x, y, \mathcal G) = g\left(\frac1{T_e}\int_{-\frac12 T_e}^{\frac12 T_e} C_i\left(x, y, P\left(t_i + t_e + (y/H - 1/2) T_{\rm ro} \right), \mathcal G\right) \,{\rm d}t_e\right),
\end{equation}
where $T_{\rm ro}$ is the rolling-shutter readout time, $T_e$ the exposure time, and $t_i$ the frame midpoint timestamp.
Following~\cite{Deblur-NeRF}, we assume a simple gamma correction model: $g(R, G, B) = (R^{1/\gamma}, G^{1/\gamma}, B^{1/\gamma})$ with $\gamma = 2.2$. As a result, the colors $\theta_j$ of the splats are defined in \emph{linear RGB} space, whereas the original version (\cref{eq:original-rendering-model}) models them directly in the gamma-corrected colour space.

We model the camera motion around the frame midpoint time $t_i$ as
\begin{equation}
  P(t_i + \Delta t) = [R\,|\,p](t) = \left[R_i \exp(\Delta t [\omega_i]_\times) \,|\, p_i + \Delta t \cdot R_i v_i\right] ,
\end{equation}
where $(v_i, \omega_i)$ are the linear and angular velocities of the camera in its local coordinate system (assumed constant throughout the frame interval $|\Delta t| < \frac12(T_e+T_{\rm ro})$).
We add the velocities as additional optimizable parameters and set their initial values to the estimate from VIO, if available, or zero otherwise.

\begin{figure}[t!]
\centering
\begin{tikzpicture}

  \node[anchor=south west,inner sep=0,minimum width=\textwidth,minimum height=.378\textwidth] (image) at (0,0) {};
  \begin{scope}[x={(image.south east)},y={(image.north west)}]

    \definecolor{arrowcolor}{HTML}{FF7F0E};
    \definecolor{splatcolor}{HTML}{1F77B4};

    \node[inner sep=0,outer sep=0] (cam) at (.8,.5) {};
    \node[inner sep=0,outer sep=0] (splat0) at (.1,.45) {};
    \node[inner sep=0,outer sep=0] (splat1) at (.38,.78) {};

    \coordinate (p0) at (.5,.48);
    \coordinate (p1) at (.6,.63);

    \draw[black!80,thick,dashed] (p0) -- (splat0);
    \draw[black!80,thick,dashed] (p1) -- (splat1);

    \draw[black!80,thick,rounded corners=1pt,draw] (.4,.15) -- (.65,.4) -- (.65,.85) -- (.4,.60) -- cycle;
    \fill[splatcolor!10!white,thick,rounded corners=1pt,opacity=.8] (.4,.15) -- (.65,.4) -- (.65,.85) -- (.4,.60) -- cycle;

    \def\splats{(.1,.45)/1.5/60,(.38,.78)/.8/-20}
    \foreach \point/\skew/\rot in \splats{
      \foreach \i/\j in {0.2107/.9,0.4463/.8,0.7133/.7,1.0217/.6,1.3863/.5,1.8326/.4,2.4079/.3,3.2189/.2,4.6052/.1} {
        \fill[opacity=\j,splatcolor,rotate around={\rot:\point}] \point ellipse ({.02*\i} and {\skew*.02*\i});
      }
      \fill[black] \point circle (2pt);
    }

    \draw[black!80,thick,dashed] (cam) -- (p0);
    \draw[black!80,thick,dashed] (cam) -- (p1);

    \draw[splatcolor,thick,rotate around={-30:(p0)}] (p0) ellipse (.015 and .1);
    \draw[splatcolor,thick,rotate around={60:(p1)}] (p1) ellipse (.01 and .05);

    \node at ($(splat0)+(0,-.25)$) {\normalsize $\mu, \Sigma$};
    \node at ($(cam)+(0,-.15)$) {\normalsize $P(t)$};
    \node at ($(.5,.48)+(.05,-.1)$) {\normalsize $\mu', \Sigma'$};

    \node at ($(cam) + (0,0)$) {\includegraphics[width=.06\textwidth]{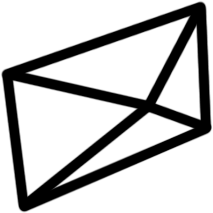}};

    \draw[->,thick,arrowcolor] (p0) -- node[black,below]{\normalsize $v'$} ++(-.07,-.1);
    \draw[->,thick,arrowcolor] (p1) -- ++(-.03,-.05);
    \draw[thick,arrowcolor] ($(cam) + (.05,.04)$) edge[->,bend right=30] ++(.05,.15);

    \node[align=left,text width=8em,font=\scriptsize] at (.8,.8) {\num{1}~A \textcolor{arrowcolor}{moving} camera causes motion blur and rolling shutter effects.};
    \node[align=left,text width=8em,font=\scriptsize] at (.6,.1) {\num{2}~We approximate the effects in the pixel coordinates.};
    \node[align=left,text width=8em,font=\scriptsize] at (.15,.85) {\num{3}~The subsequent rasterization process of the splats no longer depend on the camera pose.};

  \end{scope}
\end{tikzpicture}
\caption{Screen space approximation incorporates the motion during the frame exposure interval into our model by capturing its effect in pixel coordinates in the image plane. In our approach, the rendering model of 3DGS is decomposed into two stages: first, transforming the Gaussian parameters from world to pixel coordinates, and then rasterizing these parameters onto the image.}
\label{fig:screenspace}
\end{figure}
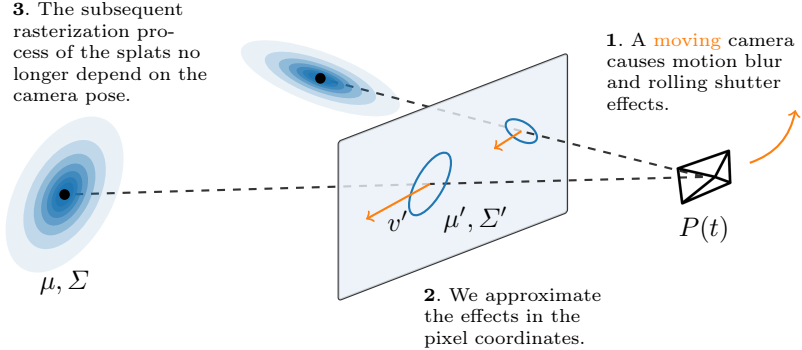

\subsection{Screen Space Approximation}
\label{sec:screen-space-approximation}
In our approach, the rendering model of 3DGS is decomposed into two stages: first, transforming the Gaussian parameters from world to pixel coordinates, and then rasterizing these parameters onto the image. This transformation is pivotal for incorporating motion effects due to camera movement during frame capture. By approximating the motion in pixel coordinates, we focus on adjusting the Gaussian means to reflect this movement, simplifying the model by primarily altering these means while keeping other parameters stable. A high-level overview of the process is presented in \cref{fig:screenspace}.

These two main phases in the 3DGS rendering model can be writen as $C_i(x, y, \mathcal G, P_i) = r(x, y, p(\mathcal G, P_i))$, where $p$ maps the Gaussian parameters $\mathcal G$ defined in world coordinates to parameters $\{ \mathcal G_{i,j}' = (\mu_{i,j}', d_{i,j}, \Sigma_{i,j}', \alpha_j, c_{i,j}) \}$ defined in pixel coordinates of the currently processed camera $i$ and individual Gaussian $j$ (see \cref{sec:gaussians-world-to-pixels} for details). The subsequent rasterization phase $r$ does not depend on the camera pose $P_i$.

We incorporate camera motion during the frame exposure interval into our model by approximating its effect in pixel coordinates as
\begin{equation}
  \mu_{i,j}'(\Delta t) \approx
  \tilde \mu_{i,j}(\Delta t) :=
  \mu_{i,j}'(0, 0) + \Delta t \cdot v_{i,j}'
\end{equation}
in particular, to simplify and optimize the implementation, we neglect the effect of (small) camera motion on the other view-dependent intermediary variables $(d_{i,j}, \Sigma_{i,j}', c_{i,j})$ and only model the effect on the pixel coordinates $\mu_{i,j}'$ of the Gaussian means and introduce a new set of variables, the \emph{pixel velocities} computed as
\begin{equation}
\label{eq:pixel-velocities}
  v_{i,j}' = -J_i (\omega_i \times \hat \mu_{i,j} + v_i),
\end{equation}
where $J_i = {\rm diag}(f_x, f_y) / d_{i,j}$ is the Jacobian of the projective transform as in \cite{GaussianSplatting} (see \cref{sec:pix-velocity-formula} for details).

\subsection{Rasterization with Pixel Velocities}
We approximate the integral in \cref{eq:motion-blur-and-rolling-shutter} with a sum of $N_{\rm blur}$ samples on a fixed uniform grid during the exposure interval
\begin{equation}
  \tilde C_i(x, y, \mathcal G) := g\left(
    \frac1{N_{\rm blur}}
    \sum_{k=1}^{N_{\rm blur}}
    \tilde C_i\left(x, y, \mathcal G, \Delta t_k(y)\right)
  \right),
  \label{eq:rendering-with-blur-and-rs-sum}
\end{equation}
where
\begin{equation}
  \Delta t_k(y) := \left(\frac{k - 1}{N_{\rm blur} - 1} - \frac12\right) \cdot T_e + \left(\frac{y}{H} - \frac12\right) \cdot T_{\rm ro}.
\end{equation}
The 3DGS alpha blending stage (\cf~\cite{GaussianSplatting}) can be written as the sum $C(x, y, \mathcal G) = \sum_j T_j c_{i,j} \alpha_{i,j}(\mu_{i,j}')$ over the depth-sorted Gaussians, where $T_j = \prod_{l=1}^{j-1}(1 - \alpha_{i,l}(\mu_{i,l}'))$ is the accumulated transmittance. The dependency on other Gaussian pixel space parameters $\mathcal G_{i,j}'$ other than $\mu_{i,j}'$ has been omitted as explained in ~\cref{sec:screen-space-approximation}. With this, \cref{eq:rendering-with-blur-and-rs-sum} becomes
\begin{equation}
  \tilde C_i(x, y, \mathcal G) := g\left(
    \frac1{N_{\rm blur}}
    \sum_{k=1}^{N_{\rm blur}}
    \sum_j T_j c_{i, j} \alpha_{i,j} (\tilde \mu_{i,j}'(\Delta t_k(y)))
  \right).
\end{equation}
Approximating camera velocity on the Gaussian means in pixel-space has the benefit that the projective transform of the 3DGS pipeline does not need to be repeated to generate each blur sample (\ie the terms in the outer sum), making the method faster than treating Gaussian velocities in world space.

\subsection{Pose Optimization}
We approximate the gradient with respect to camera pose components as
\begin{equation}
  \frac{\partial C_i}{\partial p_i} \approx -\sum_j \frac{\partial C_i}{\partial \mu_j} \quad \text{and} \quad
  \frac{\partial C_i}{\partial \nu} \approx -\sum_j \frac{\partial C_i}{\partial \mu_j} \frac{\partial R_i}{\partial \nu} R_i^\top (\mu_j - p_i),
\end{equation}
where $\nu$ is any component of $R_i$ or a parameter it depends on. This neglects the effect of small camera motions on view-dependent colours and Gaussian precisions $\Sigma_{i,j}'$ defined in pixel coordinates, but allows approximating the expression in terms of the Gaussian position derivatives $\frac{\partial C_i}{\partial \mu_{j,k}}$, which are readily available in the 3DGS pipeline.
This approach is similar to \cite{GS-SLAM} and derived in more detail in \cref{sec:camera-pose-approx-derivatives}.
To stabilize the reconstruction, we employ simple $l_2$ penalty terms to prevent the poses from drifting too far from their initial estimates.

\subsection{Evaluation Strategy - Optimization with fixed Gaussians}

Due to the well-known gauge invariance studied in classical photogrammetry (\cf~\cite{BAMS}), the inverse rendering problem has seven undetermined degrees of freedom: rotating, scaling and translating the coordinate system in all the quantities $(\mathcal G, \{ P_i, v_i, \omega_i \}_i)$ in consistent manner yields new solution $(\mathcal G', \{ P_i', v_i', \omega_i' \}_i)$ with equal photometric losses.
While, the $l_2$ penalty on the poses will fix this gauge invariance, it leaves room for small perturbations: the reconstructed scene may slightly rotate and translate, which can cause the \emph{evaluation frames} to be misaligned with reconstruction where the training frames and Gaussians $\mathcal G$ are consistent with each other.

To tackle this, we also optimize the poses and velocities of the evaluation frames with fixed gaussians $\mathcal G$, which can be achieved by including both test and training frames to the optimization problem, but disabling the back-propagation of gradient data to the parameters $\mathcal G$ (while allowing it for $P_i, v_i$ and $\omega_i$) whenever an evaluation index $i$ is selected by the SGD optimizer.

\section{Experiments}
\label{sec:experiments}
We evaluate the performance of our method on two different data sets: synthetic data based on the Deblur-NeRF data set~\cite{Deblur-NeRF} and real data recorded using mobile phones. We implemented our method as an extension to the open source 3DGS implementation \emph{Splatfacto} in Nerfstudio~\cite{Nerfstudio}, which is based on \texttt{gsplat}~\cite{gsplat} (\cf~\cref{app:experiments} for details).

\begin{table}[t!]
\centering\scriptsize
\caption{Synthetic data results comparing the baseline Splatfacto method to our approach with the appropriate compensation for each variation.\label{tab:synthetic}}
\renewcommand*{\arraystretch}{1.2}
%------------------------------------------------------------
%------------------------------------------------------------
%------------------------------------------------------------
%
%       AUTO-GENERATED FILE. Edit in generate_tables.py
%
%------------------------------------------------------------
%------------------------------------------------------------
%------------------------------------------------------------
\begin{tabular}{l|ccc|ccc|ccc|ccc}
 & \multicolumn{3}{|c}{Cozyroom} & \multicolumn{3}{|c}{Factory} & \multicolumn{3}{|c}{Pool} & \multicolumn{3}{|c}{Tanabata}\\
\hline
 & {\tiny \PSNRColName} & {\tiny \SSIMColName} & {\tiny \LPIPSColName} & {\tiny \PSNRColName} & {\tiny \SSIMColName} & {\tiny \LPIPSColName} & {\tiny \PSNRColName} & {\tiny \SSIMColName} & {\tiny \LPIPSColName} & {\tiny \PSNRColName} & {\tiny \SSIMColName} & {\tiny \LPIPSColName}\\
\hline
\multicolumn{13}{c}{\rule{0pt}{7pt}\textit{Motion blur}}\\
\hline
Baseline & 26.63 & .832 & .190 & 22.26 & .643 & .357 & \textbf{35.50} & .951 & .046 & 20.43 & .698 & .319\\
Ours & \textbf{32.20} & \textbf{.942} & \textbf{.030} & \textbf{30.67} & \textbf{.936} & \textbf{.044} & 35.27 & \textbf{.953} & \textbf{.039} & \textbf{26.71} & \textbf{.909} & \textbf{.068}\\
\hline
\multicolumn{13}{c}{\rule{0pt}{7pt}\textit{Rolling shutter effect}}\\
\hline
Baseline & 19.21 & .627 & .184 & 15.29 & .335 & .338 & 27.32 & .776 & .083 & 13.42 & .349 & .410\\
Ours & \textbf{35.84} & \textbf{.979} & \textbf{.013} & \textbf{35.27} & \textbf{.983} & \textbf{.009} & \textbf{35.11} & \textbf{.957} & \textbf{.036} & \textbf{26.03} & \textbf{.918} & \textbf{.046}\\
\hline
\multicolumn{13}{c}{\rule{0pt}{7pt}\textit{Pose noise}}\\
\hline
Baseline & 16.76 & .484 & .339 & 15.09 & .251 & .420 & 20.95 & .465 & .344 & 13.97 & .271 & .442\\
Ours & \textbf{36.30} & \textbf{.976} & \textbf{.013} & \textbf{24.03} & \textbf{.745} & \textbf{.132} & \textbf{34.68} & \textbf{.951} & \textbf{.038} & \textbf{33.53} & \textbf{.979} & \textbf{.012}\\
\hline
\end{tabular}

\end{table}

\begin{table}[t!]
  \centering\scriptsize
  \caption{Novel view synthesis results Deblur-NeRF data set variant in ~\cite{wang2023badnerf}.
  The \emph{Tanabata} scene has lower metrics due to an issue in the input data, \cf~\cref{sec:deblur-nerf-tweaks}.\label{tab:synthetic-new}}
  \renewcommand*{\arraystretch}{1.2}
  \scalebox{0.9}{
  %
  %------------------------------------------------------------
%------------------------------------------------------------
%------------------------------------------------------------
%
%       AUTO-GENERATED FILE. Edit in generate_tables.py
%
%------------------------------------------------------------
%------------------------------------------------------------
%------------------------------------------------------------
\begin{tabular}{l|cp{2.1em}p{2.5em}|cp{2.1em}p{2.5em}|cp{2.1em}p{2.5em}|cp{2.1em}p{2.5em}|cp{2.1em}p{2.5em}}
 & \multicolumn{3}{|c}{Cozyroom} & \multicolumn{3}{|c}{Factory} & \multicolumn{3}{|c}{Pool} & \multicolumn{3}{|c}{Tanabata} & \multicolumn{3}{|c}{Trolley}\\
\hline
 & {\tiny \PSNRColName} & {\tiny \SSIMColName} & {\tiny \LPIPSColName} & {\tiny \PSNRColName} & {\tiny \SSIMColName} & {\tiny \LPIPSColName} & {\tiny \PSNRColName} & {\tiny \SSIMColName} & {\tiny \LPIPSColName} & {\tiny \PSNRColName} & {\tiny \SSIMColName} & {\tiny \LPIPSColName} & {\tiny \PSNRColName} & {\tiny \SSIMColName} & {\tiny \LPIPSColName}\\
\hline
Splatfacto & 24.93 & .802 & .225 & 21.28 & .598 & .440 & 27.88 & .763 & .281 & 18.52 & .533 & .433 & 19.47 & .564 & .387\\
MPR{\tiny +}Splatf. & 29.26 & .894 & .093 & 23.38 & .737 & .246 & 30.96 & \textit{.867} & .176 & 22.77 & .773 & .227 & 26.49 & .854 & .185\\
Deblur-NeRF & 29.88 & .890 & .075 & 26.06 & .802 & .211 & 30.94 & .840 & .169 & 22.56 & .764 & .229 & 25.78 & .812 & .180\\
BAD-NeRF & \textit{30.97} & \textit{.901} & \textit{.055} & \textbf{31.65} & \textit{.904} & \textit{.123} & \textit{31.72} & .858 & \textit{.115} & \textit{23.82} & \textit{.831} & \textit{.138} & \textit{28.25} & \textit{.873} & \textit{.091}\\
Ours & \textbf{31.80} & \textbf{.945} & \textbf{.032} & \textit{30.54} & \textbf{.946} & \textbf{.078} & \textbf{32.08} & \textbf{.890} & \textbf{.075} & \textbf{24.79} & \textbf{.912} & \textbf{.079} & \textbf{30.16} & \textbf{.933} & \textbf{.044}\\
\hline
\end{tabular}

  }
\end{table}

\subsection{Synthetic Data}\label{sec:synthetic-experiments}
3DGS uses a SfM-based sparse point cloud as seed points for the initialization of the Gaussian means and scales, which is typically obtained from COLMAP~\cite{COLMAP-1}. However, with significantly blurry and other otherwise noisy data, COLMAP often fails which prevents us from obtaining good SfM point cloud for initialization on very blurry synthetic data sets like Deblur-NeRF. To overcome this, we compute an initial point cloud using SIFT~\cite{SIFT} feature matching and triangulation with the COLMAP-estimated poses.

\paragraph{BAD-NeRF data set variant} To compare our performance with existing work, we use the BAD-NeRF re-render~\cite{wang2023badnerf} of the synthetic Deblur-NeRF data set~\cite{Deblur-NeRF}. We computed the results of our method, the Splatfacto baseline and a variant where, prior to COLMAP registration, the images were deblurred with MPR~\cite{multi-stage-restoration-2021}. Similarly to~\cite{wang2023badnerf}, we only use the blurry images for pose registration in COLMAP. The results of the experiment are presented in \cref{tab:synthetic-new}, where they are also compared to the previous NeRF-based methods as reported in~\cite{wang2023badnerf}, where we also note that the original Deblur-NeRF~\cite{Deblur-NeRF} has been evaluated using ground truth training poses. Our method clearly outperforms all baselines.

\paragraph{Re-rendered data} To study the individual components of our method in more detail, we rendered new variants of the Deblur-NeRF data set to simulate motion blur, rolling shutter, and noisy pose effects respectively (see \cref{sec:deblur-nerf-tweaks} for details). We experiment on the following synthetic data set variants: {\em (i)}~a motion blur variant, which closely matches the motion blur variant of the original Deblur-NeRF data set,
{\em(ii)}~a rolling shutter only variant, {\em(iii)}~ and a translational and angular pose noise variant, without RS or motion blur effects. In the motion blur and rolling shutter variants, we use the synthetic ground truth velocities $(v_j,\omega_j)$ without additional pose noise. For these variants, we keep poses and velocities fixed during optimization. In this experiment, the ground truth poses were also used for initial point cloud triangulation instead of COLMAP-estimated poses.

We train 3DGS models for all of the above synthetic data sets using 3DGS Splatfacto method as our baseline and we compare novel-view synthesis metrics with our method with the appropriate compensation mode enabled (\ie, rolling shutter compensation but no motion blur compensation or pose optimization for RS case). The results are shown in~\cref{tab:synthetic} and qualitatively visualized in~\cref{fig:synthetic,fig:synthetic-factory}. Our method consistently outperforms the 3DGS baseline across all scenarios, indicating the effectiveness of our approach in compensating for blurring and RS effects arising from camera motion in pixel space.

\begin{table}[t]
\centering\scriptsize
\caption{PSNR metric ablation study and comparison for smartphone data.\label{tab:real}}
\setlength{\tabcolsep}{3pt}
\renewcommand*{\arraystretch}{1.2}
\scalebox{0.9}{%------------------------------------------------------------
%------------------------------------------------------------
%------------------------------------------------------------
%
%       AUTO-GENERATED FILE. Edit in generate_tables.py
%
%------------------------------------------------------------
%------------------------------------------------------------
%------------------------------------------------------------
\begin{tabular}{l|c|c|c|c|c|c|c|c}
 & \multicolumn{1}{|p{3em}}{Splatf.} & \multicolumn{1}{|p{3em}}{\hfil $\setminus$MB} & \multicolumn{1}{|p{3em}}{\hfil $\setminus$RS} & \multicolumn{1}{|p{3em}}{$\setminus$P.opt.} & \multicolumn{1}{|p{3em}}{$\setminus$V.opt.} & \multicolumn{1}{|p{3em}}{\hfil $\setminus$VIO} & \multicolumn{1}{|p{3em}}{\hfil CVR} & \multicolumn{1}{|p{3em}}{Ours}\\
\hline
Motion blur & $\textcolor{red}{-}$ & $\textcolor{red}{-}$ & \textcolor{teal}{$\checkmark$} & \textcolor{teal}{$\checkmark$} & \textcolor{teal}{$\checkmark$} & \textcolor{teal}{$\checkmark$} & \textcolor{teal}{$\checkmark$} & \textcolor{teal}{$\checkmark$}\\
Rolling shut. & $\textcolor{red}{-}$ & \textcolor{teal}{$\checkmark$} & $\textcolor{red}{-}$ & \textcolor{teal}{$\checkmark$} & \textcolor{teal}{$\checkmark$} & \textcolor{teal}{$\checkmark$} & CVR & \textcolor{teal}{$\checkmark$}\\
Pose opt. & $\textcolor{red}{-}$ & \textcolor{teal}{$\checkmark$} & \textcolor{teal}{$\checkmark$} & $\textcolor{red}{-}$ & \textcolor{teal}{$\checkmark$} & \textcolor{teal}{$\checkmark$} & \textcolor{teal}{$\checkmark$} & \textcolor{teal}{$\checkmark$}\\
Velocity opt. & $\textcolor{red}{-}$ & \textcolor{teal}{$\checkmark$} & \textcolor{teal}{$\checkmark$} & \textcolor{teal}{$\checkmark$} & $\textcolor{red}{-}$ & \textcolor{teal}{$\checkmark$} & \textcolor{teal}{$\checkmark$} & \textcolor{teal}{$\checkmark$}\\
VIO vel. init. &  & \textcolor{teal}{$\checkmark$} & \textcolor{teal}{$\checkmark$} & \textcolor{teal}{$\checkmark$} &  & $\textcolor{red}{-}$ &  & \textcolor{teal}{$\checkmark$}\\
\hline
\textsc{iphone-lego1} & 28.05 & 28.12 & \textit{29.20} & 28.59 & 28.71 & 29.03 & 23.26 & \textbf{29.20}\\
\textsc{iphone-lego2} & 27.85 & 27.88 & 27.95 & 27.39 & \textit{28.15} & \textbf{28.55} & 26.45 & 27.95\\
\textsc{iphone-lego3} & 23.75 & 23.71 & \textit{24.50} & 24.10 & 24.22 & 23.78 & 22.45 & \textbf{24.50}\\
\textsc{iphone-pots1} & 28.32 & 28.58 & 29.10 & 28.91 & 28.93 & \textbf{29.18} & 24.44 & \textit{29.10}\\
\textsc{iphone-pots2} & 27.25 & 27.39 & \textit{28.00} & 27.68 & 26.66 & 27.81 & 23.64 & \textbf{28.00}\\
\textsc{pixel5-lamp} & 28.22 & \textit{30.91} & 28.38 & 28.77 & 29.75 & 29.70 & \textbf{31.95} & 30.46\\
\textsc{pixel5-plant} & 26.57 & 27.41 & 27.45 & 26.81 & 27.37 & 27.69 & \textbf{28.20} & \textit{27.90}\\
\textsc{pixel5-table} & 28.93 & 30.82 & 29.25 & 30.33 & 31.16 & 31.47 & \textbf{32.42} & \textit{31.86}\\
\textsc{s20-bike} & 27.35 & \textit{27.74} & 27.57 & 27.58 & 27.58 & 27.72 & 26.62 & \textbf{28.93}\\
\textsc{s20-bikerack} & 25.98 & \textit{29.39} & 27.92 & 26.23 & 26.09 & 28.92 & 27.77 & \textbf{29.74}\\
\textsc{s20-sign} & 23.71 & 25.93 & 24.47 & 25.44 & 25.54 & \textit{26.28} & 24.19 & \textbf{26.84}\\
\hline
\textsc{average} & 26.91 & 27.99 & 27.62 & 27.44 & 27.65 & \textit{28.19} & 26.49 & \textbf{28.59}\\
\hline
\end{tabular}
}
\end{table}

\begin{figure}[t!]
  \centering\scriptsize

  \begin{tikzpicture}[inner sep=0]

    \setlength{\figurewidth}{.24\textwidth}
    \setlength{\figureheight}{.75\figurewidth}
    \setlength{\figurerowheight}{-1.02\figureheight}

    \node[anchor=center, rotate=90] (label-top) at (-1em,0.5\figurerowheight) {\bf Baseline};
    \node[anchor=center, rotate=90] (label-top) at (-1em,1.5\figurerowheight) {\bf Ours};
    \node[anchor=center, rotate=90] (label-top) at (-1em,2.5\figurerowheight) {\bf Reference};
    \node[anchor=center, rotate=90] (label-top) at (-1em,3.5\figurerowheight) {\bf $l_2$ contrib.};
    \node[anchor=center, rotate=90] (label-top) at (-1em,4.5\figurerowheight) {\bf $\Delta l_2$};

    \foreach \imgtype/\row in {
      baseline/0,
      pose_opt-motion_blur-rolling_shutter-velocity_opt/1,
      gt/2,
      0_l2-ours/3,
      0_l2-diff/4} {
      \foreach \file/\label/\cols [count=\col from 0] in {
          colmap-sai-cli-vels-blur-scored-iphone-lego1/{lego1/iPhone}/0,
          colmap-sai-cli-vels-blur-scored-iphone-pots2//1,
          colmap-sai-cli-vels-blur-scored-pixel5-table//2,
          colmap-sai-cli-vels-blur-scored-s20-bike//3
        } {
          \begin{scope}

          \node[anchor=north west, inner sep=2pt] (label-\row-\col) at (\col*\figurewidth,\row*\figurerowheight) {\label};

          \iflabel{\label}{\col}{%
            \clip[rounded corners=2pt] (\col*\figurewidth,\figurerowheight*\row) -- (\col*\figurewidth+.99\figurewidth,\figurerowheight*\row) -- (\col*\figurewidth+.99\figurewidth,\figurerowheight*\row-\figureheight) -- (\col*\figurewidth,\figurerowheight*\row-\figureheight) -- cycle;
          }{%
            \clip[rounded corners=2pt] (label-\row-\col.south west) -- (label-\row-\col.south east) -- (label-\row-\col.north east) -- (\col*\figurewidth+.99\figurewidth,\figurerowheight*\row) -- (\col*\figurewidth+.99\figurewidth,\figurerowheight*\row-\figureheight) -- (\col*\figurewidth,\figurerowheight*\row-\figureheight) -- cycle;
          }

          \node [inner sep=0,minimum width=\figurewidth,minimum height=\figureheight,fill=red!10!white,anchor=north west] at (\col*\figurewidth, \row*\figurerowheight) {%
              \includegraphics[height=\figureheight]{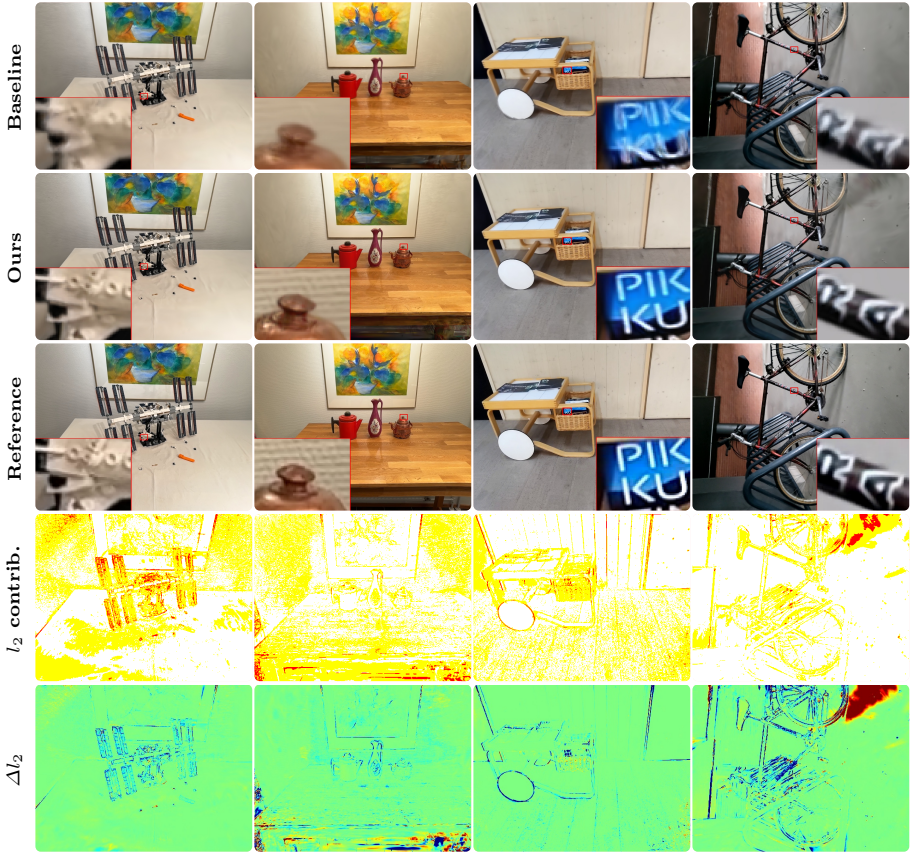}%
          };

          \end{scope}
      }
    }

  \end{tikzpicture}
  \caption{Real data examples with scenes captured with smartphones. From top to bottom:
  Using COLMAP poses without motion blur compensation (baseline);
  with motion blur compensation (ours);
  reference evaluation image;
  $l_2$ error contributions, ours (red: 30\%, yellow: 60\%, white: 10\%);
  $l_2$ error differences ours vs baseline (red: more error, blue: less error). Scenes: \texttt{lego1} (iPhone), \texttt{pots2} (iPhone), \texttt{table} (Pixel), \texttt{bike} (S20).\looseness-1}
  \label{fig:real}
\end{figure}

\begin{figure}[t!]
  \centering\scriptsize

  \begin{tikzpicture}[inner sep=0]

    \setlength{\figurewidth}{.24\textwidth}
    \setlength{\figureheight}{.75\figurewidth}
    \setlength{\figurerowheight}{-1.02\figureheight}

    \node[anchor=center, rotate=90] (label-top) at (-1em,0.5\figurerowheight) {\bf Baseline};
    \node[anchor=center, rotate=90] (label-top) at (-1em,1.5\figurerowheight) {\bf Ours (MB)};
    \node[anchor=center, rotate=90] (label-top) at (-1em,2.5\figurerowheight) {\bf Ours (MBRS)};
    \node[anchor=center, rotate=90] (label-top) at (-1em,3.5\figurerowheight) {\bf Reference};

    \foreach \imgtype/\row in {
      baseline/0,
      pose_opt-motion_blur-velocity_opt/1,
      pose_opt-motion_blur-rolling_shutter-velocity_opt/2,
      gt/3} {
      \foreach \file/\label/\col in {
          colmap-sai-cli-vels-blur-scored-pixel5-table-rsabl//0,
          colmap-sai-cli-vels-blur-scored-pixel5-lamp-rsabl/1,
          colmap-sai-cli-vels-blur-scored-s20-sign-rsabl//2,
          colmap-sai-cli-vels-blur-scored-s20-bike-rsabl//3} {
          \begin{scope}

          \node[anchor=north west, inner sep=2pt] (label-\row-\col) at (\col*\figurewidth,\row*\figurerowheight) {\label};
          \clip[rounded corners=2pt] (\col*\figurewidth,\figurerowheight*\row) -- (\col*\figurewidth+.99\figurewidth,\figurerowheight*\row) -- (\col*\figurewidth+.99\figurewidth,\figurerowheight*\row-\figureheight) -- (\col*\figurewidth,\figurerowheight*\row-\figureheight) -- cycle;

          \node [inner sep=0,minimum width=\figurewidth,minimum height=\figureheight,fill=red!10!white,anchor=north west] at (\col*\figurewidth, \row*\figurerowheight) {%
              \includegraphics[height=\figureheight]{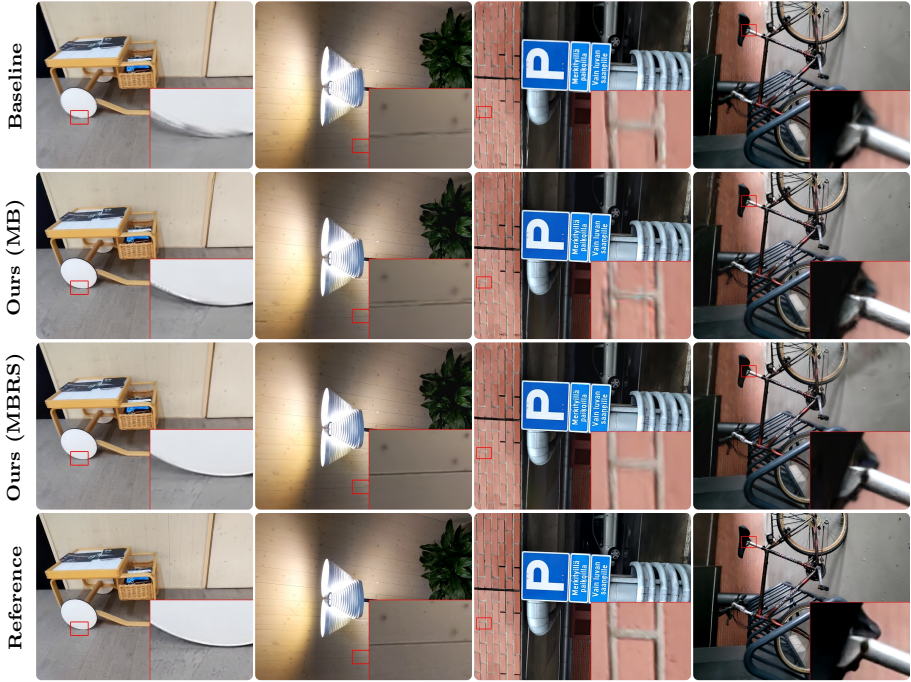}%
          };

          \end{scope}
      }
    }

  \end{tikzpicture}
  \caption{Empirical ablations on deblurring with and without rolling shutter pose optimization on four real-world smartphone data scenes with rolling shutter effects. Our method with both motion blur and rolling shutter compensation (MBRS) gives sharper artefact-free reconstructions compared to just motion blur compensation (MB).}
  \label{fig:deblurring-with-and-without-rs-pose-opt}
\end{figure}

\subsection{Smartphone Data}\label{sec:real-data}
To the end of real data evaluation, we use a new data set recorded using three different smartphones: Samsung S20 FE, Google Pixel 5 and iPhone 15 Pro. The first two are Android phones with a known, and relatively large rolling-shutter readout time $T_{\rm ro}$. %

The data set consists of 11 short handheld recordings of various scenes collected using the Spectacular~Rec application~\cite{SpectacularAIMappingTools}, which records synchronized IMU and video data, together with the built-in (factory) calibration information from each device. The same application is available both for Android and iOS,  and it was used to capture the raw image and IMU data on the devices.

\paragraph{Preprocessing and VIO velocity estimation} We first process the recorded data with the Spectacular AI SDK~\cite{SpectacularAIMappingTools}, to obtain the following: {\em{(i)}}~A sparse set of key frames with minimum distance of 10~cm, selected to approximately minimize motion blur among the set of candidate key frames (see \cref{sec:key-frame-selection} for details). {\em{(ii)}}~For each key frame, the initial estimates for the frame velocities $(v_j^{\rm VIO},\omega_j^{\rm VIO})$, based on the fusion of IMU and video data. {\em{(iii)}}~Approximate VISLAM-based poses $P_j^{\rm SAI}$.

\paragraph{Training and evaluation split} As with the synthetic data and in \cite{Deblur-NeRF}, we aim to select the least blurry frames for evaluation, which is performed by splitting the (ordered) key frames to subsets of eight consecutive key frames, and for each subset, picking the least blurry one for evaluation. We use the same motion blur metric as for key frame selection in \cref{sec:key-frame-selection}.

\paragraph{Pose and intrinsic estimation} After preprocessing, the poses of the key frames and camera intrinsics are estimated using COLMAP~\cite{COLMAP-1} (through Nerfstudio). In our main results, we only included sequences in which COLMAP did not fail due to excessively difficult visual conditions. %
In \cref{sec:additional-smartphone-experiments} we present additional experiments where manual calibration results or built-in calibration data from the devices is used in place of COLMAP intrinsics.

\paragraph{COLMAP baseline} Our baseline comparison involves the Splatfacto method paired with poses estimated by COLMAP, without any specific compensation for motion blur or rolling shutter effects. This baseline serves as a reference point for evaluating the efficacy of our proposed motion blur compensation method.

\paragraph{CVR de-rolling baseline} As a baseline for rolling shutter compensation, we experimented with a variant where, prior to the COLMAP phase, the image data was de-rolled with CVR~\cite{CVR-2022}. For a fair comparison, we enabled the other optimization features in our method in this test.
We also note that this test was omitted with synthetic data as CVR requires access to consecutive frames, which were not available in the sparse synthetic test. %

\paragraph{Ablation study} We also performed an ablation study where each of the main components of the method are individually disabled: {\em{(i)}}~motion blur compensation, by setting $T_e = 0$, $N_{\rm blur} = 1$; {\em{(ii)}}~rolling shutter compensation by setting $T_{\rm ro} = 0$; {\em{(iii)}}~pose optimization; {\em{(iv)}}~pptimization of linear and angular velocities $(v_i, \omega_i)$; {\em{(v)}}~velocity optimization initialization from VIO by setting $v_i = \omega_i = 0$ instead of $v_i = v_i^{\rm VIO}$, $\omega_i = \omega_i^{\rm VIO}$.

\paragraph{Results} The real data results are given in \cref{tab:real}. All features included in the ablation study display a positive impact and our method achieves the best overall performance in therms of the PSNR metric.
Performance compared to CVR pre-processing is mixed: while CVR combined with our other compensation features performs best on the data from Pixel 5, it is worse on S20 and clearly worse if enabled for iPhone (with low rolling shutter readout time). Qualitatively we noticed that CVR outputs are often contaminated by various types of artefacts that cause bad performance in 3DGS, and conclude that our method is more robust than CVR for rolling-shutter compensation for static scene reconstructions.

Qualitatively, the sharpness of the reconstructions is subtly, but noticeably increased when motion blur compensation is enabled, as demonstrated in the highlighted details in \cref{fig:real,fig:real2}. Furthermore, \cref{fig:deblurring-with-and-without-rs-pose-opt} shows that rolling shutter compensation also noticeably improves the reconstruction quality, especially closer to the edges of the evaluation images. 

The automatic selection of key frames for evaluation also played a significant role in our analysis, particularly in highlighting areas poorly represented in the training data. These areas, especially near the edges of the visual field, were more susceptible to artefacts that disproportionately affected the PSNR metric, underlining the challenges in balancing motion blur compensation with the preservation of image quality across the entire scene. These effects are visible near the edges of the error metric figures in~\cref{fig:real,fig:real2}.

\begin{table}[t]
\centering\scriptsize
\caption{Real data timing results. Training wall clock time $T$ (minutes) in the baseline method and ours with different features disabled.\label{tab:real-meta}}
\setlength{\tabcolsep}{10pt}
\renewcommand*{\arraystretch}{1.2}
%------------------------------------------------------------
%------------------------------------------------------------
%------------------------------------------------------------
%
%       AUTO-GENERATED FILE. Edit in generate_tables.py
%
%------------------------------------------------------------
%------------------------------------------------------------
%------------------------------------------------------------
\begin{tabular}{l|c|c|c|c|c|c}
 &  &  & \multicolumn{1}{|c}{Splatfacto} & \multicolumn{1}{|c}{\hfil $\setminus$MB} & \multicolumn{1}{|c}{\hfil $\setminus$RS} & \multicolumn{1}{|c}{Ours}\\
\hline
\textsc{iphone-lego1} & 21 & 1920$\times$1440 & 19 & 21 & 81 & 81\\
\textsc{iphone-lego2} & 23 & 1920$\times$1440 & 19 & 22 & 78 & 78\\
\textsc{iphone-lego3} & 21 & 1920$\times$1440 & 22 & 28 & 114 & 114\\
\textsc{iphone-pots1} & 25 & 1920$\times$1440 & 19 & 22 & 76 & 76\\
\textsc{iphone-pots2} & 26 & 1920$\times$1440 & 20 & 23 & 83 & 83\\
\textsc{pixel5-lamp} & 35 & 1600$\times$1200 & 16 & 23 & 62 & 67\\
\textsc{pixel5-plant} & 30 & 1600$\times$1200 & 15 & 19 & 52 & 58\\
\textsc{pixel5-table} & 23 & 1600$\times$1200 & 16 & 20 & 55 & 62\\
\textsc{s20-bike} & 43 & 1920$\times$1440 & 21 & 26 & 79 & 90\\
\textsc{s20-bikerack} & 32 & 1920$\times$1440 & 20 & 26 & 71 & 79\\
\textsc{s20-sign} & 28 & 1920$\times$1440 & 21 & 36 & 162 & 210\\
\hline
\textsc{average} &  &  & 19 & 24 & 83 & 91\\
\hline
\end{tabular}

\end{table}

\paragraph{Timing tests} The training times corresponding to selected experiments in~\cref{tab:real} (which were computed on NVidia A100 GPUs) are shown in~\cref{tab:real-meta}. The increase in training times (here shown for a single test run) varies significantly by case. On the average, our full method with $N_{\rm blur} = 5$ samples is approximately five times slower than the baseline. Disabling rolling shutter compensation only has a small, and possibly random effect. However, disabling the motion blur compensation shows that our rolling shutter has a low overhead (26\%) compared to the baseline. 
We also note that our CUDA implementation was not yet profiled for new bottlenecks, and could have significant potential for further speed improvement for motion blur compensation.
The memory consumption of our approach is not significantly higher than for the baseline Splatfacto model. For development purposes, we also successfully trained on consumer grade GPUs, such as NVidia RTX 4060 TI with 16~GB of VRAM.

\section{Discussion and Conclusion}
\label{sec:discussion}
We demonstrated how blurring and rolling shutter effects can be efficiently implemented in the Gaussian Splatting (3DGS) framework. We quantitatively and qualitatively demonstrated that the method improves over the 3DGS Splatfacto baseline in both synthetic and real data experiments. Furthermore, we showed that our method outperforms the previous NeRF-based state-of-the-art, BAD-NeRF~\cite{wang2023badnerf} on synthetic data. We also demonstrated superior performance and robustness compared to baselines where the data was first deblurred or de-rolled using modern deep-learning-based methods, MPR~\cite{multi-stage-restoration-2021} or CVR~\cite{CVR-2022}.

Incorporating learning-based aspects directly into the 3D model generation instead of the 2D input data is a new and promising approach studied in, \eg, \cite{weber2023nerfiller,wu2023reconfusion}. We believe that, in the context of differentiable rendering and 3D reconstruction, this approach is likely to prove superior to learning-based image manipulation as a pre-processing step.
Modelling local linear trajectories with more complex spline-based shapes as in \cite{wang2023badnerf, li2024usbnerf} could be an additional follow up work.\looseness-1

Data capture in-the-wild is typically done on smart phones with a rolling shutter sensor and during relative motion. These effects are rarely represented well in modern benchmark data for inverse rendering, limiting the real-world use of 3DGS based methods. Our work represents a significant step forward in the integration of motion blur and rolling shutter corrections within the 3DGS framework, opening up new avenues for research and application in differentiable rendering and 3D scene reconstruction.

\codepublicationnote

\section*{Acknowledgements} 
MT was supported by the Research Council of Finland Flagship programme: Finnish Center for Artificial Intelligence (FCAI). AS acknowledges funding from the Research Council of Finland (grant id 339730). We acknowledge CSC -- IT Center for Science, Finland, for computational resources.

\clearpage

\bibliographystyle{splncs04}

\clearpage
\appendix
\setcounter{page}{1}
{
\centering
\Large
\vspace*{0.5em}\textbf{Supplementary Material} \\
\vspace{1.0em}
}

\section{Method Details}
\label{app:methods}

\subsection{Gaussian parametrization}
\label{sec:gaussians-parametrization}

In \texttt{gsplat}, the Gaussian covariances $\Sigma$ in world coordinates are parametrized as
\begin{equation}
  \Sigma = R(q) \, {\rm diag}({\rm sigmoid}(s_1, s_2, s_3))\, R(q)^\top
\end{equation}
where $q \in \mathbb{R}^4$ is a normalized quaternion corresponding to a rotation matrix $R(q) \in \mathrm{SO}(3)$ and $(s_1,s_2,s_3) \in \mathbb{R}$ are scale parameters.
This is not changed in our implementation, but we use $\Sigma$ instead of the above parameters in this paper for brevity and clarity.

\subsection{Transforming Gaussians from world to pixel coordinates}
\label{sec:gaussians-world-to-pixels}

In 3DGS, the rendering equation can be written as,
\begin{equation}
\label{eq:rendering-equation}
  C_i(x, y, P, \mathcal G) = r(x, y, \pi(\hat p(\mathcal G, P)))
\end{equation}
where $\hat p : (\mu, \Sigma, \theta, P) \mapsto (\hat \mu, \hat \Sigma, c)$ maps each (visible) Gaussian from the world coordinate system to the \emph{camera coordinate} system
\begin{equation}
\label{eq:project-gaussians-equations}
  \hat \mu_{i,j} = R_i^\top (\mu_j - p_i), \quad
  \hat \Sigma_{i,j} = R_i^\top \Sigma_j R_i, \quad
  c_{i,j} = \theta_j(\frac{\mu_j - p_i}{\Vert{\mu_j - p_i\Vert}})
\end{equation}
where the colour $c$ is computed by evaluation the spherical harmonic function at the normalized viewing direction. We use spherical harmonics of degree three. The function $\pi : (\hat \mu, \hat \Sigma) \mapsto (\mu', d, \Sigma')$ projects Gaussians from camera coordinates to pixel coordinates with depth:
\begin{equation}
  d = \tilde \mu_z, \quad
  \mu' = (\tilde \mu_x / d, \tilde \mu_y / d), \quad
  \Sigma' = J_i \hat \Sigma J_i^\top,
\end{equation}
where $\tilde \mu = K_i \hat \mu$ and $J_i = J' K_i$
is the Jacobian matrix of the pinhole camera projection $\hat \mu \mapsto \mu'$ with intrisic camera matrix $K_i$: %
\begin{equation}
\label{eq:pinhole-projection-jacobian}
  J' = \frac1{d}
  \left[
    \begin{matrix}
      1 & \;0 &\; -\tilde \mu_x/d \\
      0 & \;1 &\; -\tilde \mu_y/d \\
    \end{matrix}
  \right],
  \quad
  K_i = \left[
    \begin{matrix}
      f_x & 0 & c_x \\
      0 & f_y & c_y \\
      0 & 0 & 1
    \end{matrix}
  \right].
\end{equation}
Note that unlike the original \texttt{gsplat} \cite{gsplat-2023mathematical} and the Inria implementations, we do not use the OpenGL NDC coordinate system as an intermediate step between projecting Gaussians to pixel coordinates.

The Gaussian with low depth $d < d_{\min}$ or pixel coordinates $(\mu'_x, \mu'_y)$ too far outside the image boundaries $[0,W) \times [0,H)$ are discarded in the next rendering phases represented by the function $r$.

\subsection{Differentiation with respect to the camera pose}
\label{sec:camera-pose-approx-derivatives}

We seek to differentiate the rendering equation \cref{eq:rendering-equation} with respect to the current camera pose parameters $P_i \in \mathrm{SE}(3)$. The key to camera pose optimization is differentiating the intermediate projection terms \cref{eq:project-gaussians-equations} with respect to some parametrization of $P_i = [R_i\,|\,p_i]$. First, note that the Jacobian\footnote{Slight abuse of notation: we use $\frac{\partial}{\partial \nu}$ notation for all derivatives, including Jacobian matrices and multi-dimensional tensors, such as $\frac{\partial \hat \Sigma_{i,j}}{\partial p_i}$} of the Gaussian mean in camera coordinates with respect to the camera center $p_i$ is
\begin{equation}
\label{eq:derivative-pixel-mean-pose}
  \frac{\partial \hat \mu_{i,j}}{\partial p_i} = -R_i^\top = -\frac{\partial \hat \mu_{i,j}}{\partial \mu_j}
\end{equation}
and if we neglect the effect of $p_i$ on $c_{i,j}$, \ie, the view-dependency of the colors, and on the covariances $\hat \Sigma_{i,j}$, we can approximate:
\begin{align*}
  \frac{\partial C_i}{\partial p_i}
  &= \sum_j \left(\frac{\partial C_i}{\partial \hat \mu_{i,j}} \frac{\partial \hat \mu_{i,j}}{\partial p_i}
  + \frac{\partial C_i}{\partial \hat \Sigma_{i,j}} \frac{\partial \hat \Sigma_{i,j}}{\partial p_i}
  + \frac{\partial C_i}{\partial c_{i,j}} \frac{\partial c_{i,j}}{\partial p_i}
  \right) \\
  &\approx \sum_j \frac{\partial C_i}{\partial \hat \mu_{i,j}} \frac{\partial \hat \mu_{i,j}}{\partial p_i}
  = -\sum_j \frac{\partial C_i}{\partial \hat \mu_{i,j}} \frac{\partial \hat \mu_{i,j}}{\mu_j}
  \approx -\sum_j \frac{\partial C_i}{\partial \mu_j},
\end{align*}
that is, moving the camera to direction $\Delta p$ is approximately the same as moving all the visible Gaussians to the opposite direction $-\Delta p$.

Furthermore, using \cref{eq:derivative-pixel-mean-pose} and ignoring effects on color-dependency, we can use the following expression
\begin{equation}
  \frac{\partial C_i}{\partial \mu_j}
  =  \frac{\partial C_i}{\partial \hat \mu_{i,j}} \frac{\partial \hat \mu_{i,j}}{\partial \mu_j} +  \frac{\partial C_i}{\partial c_{i,j}} \frac{\partial c_{i,j}}{\partial \mu_j}
  \approx \frac{\partial C_i}{\partial \hat \mu_{i,j}} \frac{\partial \hat \mu_{i,j}}{\partial \mu_j}
  = \frac{\partial C_i}{\partial \hat \mu_{i,j}} R_i^\top
\end{equation}
The derivative of the rendering equation with respect to the rotation of the camera pose can also be expressed by neglecting the effect of small rotations on the shape of $\hat \Sigma_{i,j}$ of the Gaussians in camera coordinates. The derivative of the pixel colour $C_i$ with respect to any rotation matrix $R_i$ component $\nu$ becomes

\begin{align*}
  \frac{\partial C_i}{\partial \nu}
  &\approx \sum_j \frac{\partial C_i}{\partial \hat \mu_{i,j}} \frac{\partial \hat \mu_{i,j}}{\partial \nu}
  \approx \sum_j \frac{\partial C_i}{\partial \mu_j} R_i \frac{\partial \hat \mu_{i,j}}{\partial \nu}
  = \sum_j \frac{\partial C_i}{\partial \mu_j} R_i \left(\frac{\partial R_i}{\partial \nu}\right)^\top (\mu_j - p_i) \\
  &= \sum_j \frac{\partial C_i}{\partial \mu_j} R_i \left(\frac{\partial R_i}{\partial \nu}\right)^\top R_i \hat \mu_{i,j}
  = -\sum_j \frac{\partial C_i}{\partial \mu_j} \frac{\partial R_i}{\partial \nu} \hat \mu_{i,j}
\end{align*}
The benefit of this approach is that derivatives with respect to both rotation $R_i$ and translation $p_i$ of the camera can be approximated in terms of the derivatives $\frac{\partial C_i}{\partial \mu_j}$ with respect to the Gaussian means, and other properties that are readily available in the 3DGS backwards pass.

The above formulas hold for camera-to-world transformations $P = [R | p]$. For their world-to-camera counterparts $P' = [R' | p'] = [R^\top | -R^\top p]$, we can first change \cref{eq:project-gaussians-equations} to $\hat \mu_{i,j} = R_i' \mu_j + p'$ and then write
\begin{equation}
  \frac{\partial C_i}{\partial \mu_j}
  \approx \frac{\partial C_i}{\partial \hat \mu_{i,j}} \frac{\partial \hat \mu_{i,j}}{\partial \mu_j}
  = \frac{\partial C_i}{\partial \hat \mu_{i,j}} R_i',
  \quad
  \frac{\partial \hat \mu_{i,j}}{\partial p_i'} = I
  = R_i' (R_i')^\top =  \frac{\partial \hat \mu_{i,j}}{\partial \mu_j} (R_i')^\top
\end{equation}
and, consequently
\begin{equation}
\label{eq:approx-cam-trans-derivative-w2c}
  \frac{\partial C_i}{\partial p_i}
  \approx \sum_j \frac{\partial C_i}{\partial \hat \mu_{i,j}} \frac{\partial \hat \mu_{i,j}}{\partial p_i'}
  = \sum_j \frac{\partial C_i}{\partial \hat \mu_{i,j}} \frac{\partial \hat \mu_{i,j}}{\mu_j} (R_i')^\top
  \approx \sum_j \frac{\partial C_i}{\partial\mu_j} (R_i')^\top
\end{equation}
and
\begin{equation}
  \label{eq:approx-cam-rot-derivative-w2c}
  \frac{\partial C_i}{\partial \nu}
  \approx \sum_j \frac{\partial C_i}{\partial \mu_j} (R_i')^\top \frac{\partial R_i'}{\partial \nu} \mu_j.
\end{equation}
The latter format in \cref{eq:approx-cam-trans-derivative-w2c} and \cref{eq:approx-cam-rot-derivative-w2c} is used in our implementation.

\subsection{Derivation of the pixel velocity formula}
\label{sec:pix-velocity-formula}

The derivative of the pixel coordinates $\mu'$ of a Gaussian center with respect to camera motion $P(t)$ is
\begin{align*}
  \frac{\rm d}{\rm d t} \mu'(P(t))
    &= J_{\hat \mu \mapsto \mu', t} \frac{\rm d}{\rm d t} \hat \mu(P(t)) \\
    &= J_i \frac{\rm d}{\rm d t} R_i^\top (t) (\mu_j - p_i(t)) \\
    &= J_i \frac{\rm d}{\rm d t} (R_i \exp(t [\omega_j]_\times))^\top (\mu_j - (p_i + t \cdot R_i v_i)) \\
    &= J_i (([\omega_j]_\times^\top \exp(t [\omega_j]_\times)^\top R_i^\top)(\mu_j - p_i(t)) - \exp(t [\omega_j]_\times)^\top R_i^\top (R_i v_i)) \\
    &= J_i ([\omega_j]_\times^\top \exp(t [\omega_j]_\times)^\top \hat \mu_{i,j} - \exp(t [\omega_j]_\times)^\top v_i).
\end{align*}
The derivative at $t = 0$ is
\begin{equation*}
  \frac{\rm d}{\rm d t} \mu'(P(t))|_{t=0} = J_i ([\omega_j]_\times^\top  \hat \mu_{i,j} - v_i)
  = - J_i (\omega_j \times \hat \mu_{i,j} + v_i) = v_{i,j}'.
\end{equation*}

\subsection{Key Frame Selection}
\label{sec:key-frame-selection}
The level of motion blur in a given frame can be estimated using a VIO or VISLAM system by examining the pixel velocities of the 3D positions $l_j$, $j = 1,\ldots,N^{\rm lm}_i$ of the sparse SLAM landmarks visible in the frame in question. The Open Source part of the Spectacular AI Mapping tools~\cite{SpectacularAIMappingTools} computes a \emph{motion blur score} as
\begin{equation}
\label{eq:motion-blur-score}
  M_i = \frac{1}{N^{\rm lm}_i} \sum_j \left \| J_i
    \left(
      \omega_i \times (R_i\, l_j + p_i) + v_i
    \right) \right\|
\end{equation}
where $J_i$ is as in \cref{eq:pinhole-projection-jacobian}, $(R_i, t_i)$ is the world-to-camera pose and $(v_i, \omega_i)$ are the instantaneous camera-coordinate linear and angular velocities of the frame, respectively. %

To reduce sporadic motion blur, the Spectacular AI software drops all key frame candidates $i$ which have the highest blur score in a neighbourhood of 4 key frame candidates ($[i-2,i-1,i,i+1]$). We utilize the motion blur score computed using \cref{eq:motion-blur-score} also for the training and evaluation subset partitions described in \cref{sec:real-data}.

Additionally, we ran an experiment where the evaluation frames were picked by selecting every 8th key frame and switched off the motion blur score based filtering. See \cref{tab:without-motion-keyframe-filter}. The numeric results were very similar with motion blur filtering enabled. In the main paper, we chose to include the results with motion blur filtering enabled, to demonstrate that our approach can decrease motion-blur-induced effects even after the easy options for filtering blurry input frames have been exhausted. This highlights the usefulness of deblurring utilizing a 3D image formation model in the context of differentiable rendering.

\subsection{Transferring velocities from one SLAM method to another}

Assuming we have a matching set of $N$ poses $P_i = (R_i, p_i)$ for two methods (COLMAP and SAI), which differ by a ${\rm Sim}(3)$ transformation, we can map the linear frame velocities as
\begin{equation}
\label{eq:velocity-scaling-colmap-sai}
  v_i = \frac{s(p^{\rm COLMAP})}{s(p^{\rm SAI})} v_i^{\rm SAI}, \quad s^2(p) := \sum_i \left\|  p_i - \frac{\sum_i p_i}{N} \right\|^2.
\end{equation}
Note that the result does not depend on the rotation or translation part of the transformation and the accuracy of the result depends on the average scale consistency of each method.

\section{Data Sets and Metrics}
\label{sec:deblur-nerf-tweaks}

\paragraph{Modifications to the Deblur-NeRF data set}
The original version of the Deblur-NeRF data set \cite{Deblur-NeRF} (or the BAD-NeRF re-render~\cite{wang2023badnerf}) did not include the velocity data, which was randomly generated by the original authors with an unspecified seed number. We regenerated the images using the Blender \cite{Blender} files, which we modified by fixing the random seed, and also included velocity information in our outputs. All of the models mentioned in the \cite{Deblur-NeRF} were also not available (\eg `trolley'), and those cases are omitted from our version of the data set.

As in~\cite{Deblur-NeRF}, the evaluation images in the data set were rendered without blur or rollings shutter effects, which allows us to assess the novel-view synthesis performance of the underlying `sharp' reconstruction independent of the accuracy of our simulation and compensation for these effects.

Furthermore, we switched from the custom 10-sample blur implementation to Blender's built-in motion blur and rolling-shutter effects, which are less prone to sampling artefacts. In addition, we adjusted the caustics rendering settings in the Blender scenes to reduce non-deterministic raytracing sampling noise in the data with reasonable sample counts. As a result, our modified data set, published as supplementary material, provides effectively deterministic rendering capabilities and improves the reproducibility of the results compared to the original version.
Our modified data set also includes de-focus blur, which was studied in~\cite{Deblur-NeRF} but not in this paper, which is why this variation is also omitted from~\cref{tab:synthetic}.

We also note that the different versions of the Deblur-NeRF data set have different gamma correction factors. The original version~\cite{Deblur-NeRF} and our re-render use $\gamma=2.2$ while the BAD-NeRF~\cite{wang2023badnerf} re-render use $\gamma=1$. In \cref{sec:synthetic-experiments}, we modify the $\gamma$ parameter of our method accordingly.

Finally, we observed that the \emph{Tanabata} scene in the BAD-NeRF re-render contains an error that manifests as lower metrics: the sharp evaluation images and blurry training images have been rendered using a slightly different 3D model, apparently unintentionally. This issue is not present in the original version in \cite{Deblur-NeRF}. 

\section{Experiment Details}
\label{app:experiments}

\paragraph{Splatfacto hyperparameters}
The baseline method in our tests is Splatfacto in Nerfstudio version 1.1.0, which includes several improvements from the recent works \cite{ye2024absgs,Yu2023MipSplatting,xie2023physgaussian}. For the \texttt{gsplat} code, we used the main branch version accessed on 2024-05-07
and our method has also been implemented based on these versions.

The following non-default parameters were used for both the Splatfacto baseline and our method: {\em{(i)}}~limited iteration count to 20k. {\em{(ii)}}~enabled antialiased rasterization; (\cf~\cite{Yu2023MipSplatting}); {\em{(iii)}}~enabled scale regularization (\cf~\cite{xie2023physgaussian}).
Furthermore, for smartphone data where our method was used with $\gamma=2.2$, the minimum RGB level of all color channels was set to 10 in the training data, to avoid large negative logit color values, which resulted in artefacts in the proximity of dark areas in certain scenes. This modification was only used in our method and not the baseline Splatfacto.

\section{Additional Results}\label{sec:additional-smartphone-experiments}
\begin{figure}[b]
  \centering\scriptsize

  \begin{tikzpicture}[inner sep=0]

    \setlength{\figurewidth}{.3\textwidth}
    \setlength{\figureheight}{.75\figurewidth}
    \setlength{\figurerowheight}{-1.02\figureheight}

    \node[anchor=center, rotate=90] (label-top) at (-1em,0.5\figurerowheight) {\bf Baseline};
    \node[anchor=center, rotate=90] (label-top) at (-1em,1.5\figurerowheight) {\bf Ours};

    \foreach \file/\label/\row/\col [count=\i] in {
        synthetic-tanabata-mb-baseline/Motion blur/0/0,
        synthetic-tanabata-rs-baseline/Rolling shutter/0/1,
        synthetic-tanabata-posenoise-baseline/Pose noise/0/2,
        synthetic-tanabata-mb-motion_blur/Motion blur/1/0,
        synthetic-tanabata-rs-rolling_shutter/Rolling shutter/1/1,
        synthetic-tanabata-posenoise-pose_opt/Pose noise/1/2
      } {
        \begin{scope}

        \node[anchor=north west, inner sep=2pt] (label-\i) at (\col*\figurewidth,\row*\figurerowheight) {\tiny\vphantom{g}\label};
        \clip[rounded corners=2pt] (label-\i.south west) -- (label-\i.south east) -- (label-\i.north east) -- (\col*\figurewidth+.99\figurewidth,\figurerowheight*\row) -- (\col*\figurewidth+.99\figurewidth,\figurerowheight*\row-\figureheight) -- (\col*\figurewidth,\figurerowheight*\row-\figureheight) -- cycle;

        \node [inner sep=0,minimum width=\figurewidth,minimum height=\figureheight,fill=red!10!white,anchor=north west] at (\col*\figurewidth, \row*\figurerowheight) {%
            \includegraphics[height=\figureheight]{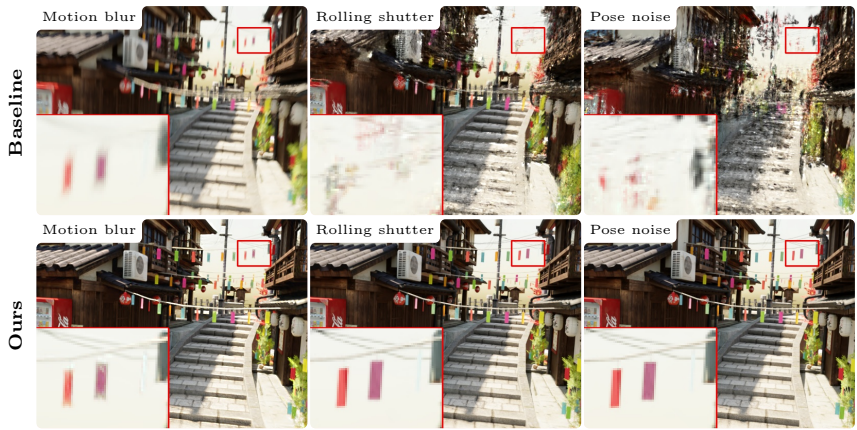}%
        };

        \end{scope}
    }

  \end{tikzpicture}
  \caption{3DGS reconstructions from the synthetic \texttt{tanabata} scene}
  \label{fig:synthetic-tanabata}
\end{figure}

\begin{figure}[b]
  \centering\scriptsize

  \begin{tikzpicture}[inner sep=0]

    \setlength{\figurewidth}{.3\textwidth}
    \setlength{\figureheight}{.75\figurewidth}
    \setlength{\figurerowheight}{-1.02\figureheight}

    \node[anchor=center, rotate=90] (label-top) at (-1em,0.5\figurerowheight) {\bf Baseline};
    \node[anchor=center, rotate=90] (label-top) at (-1em,1.5\figurerowheight) {\bf Ours};

    \foreach \file/\label/\row/\col [count=\i] in {
        synthetic-pool-mb-baseline/Motion blur/0/0,
        synthetic-pool-rs-baseline/Rolling shutter/0/1,
        synthetic-pool-posenoise-baseline/Pose noise/0/2,
        synthetic-pool-mb-motion_blur/Motion blur/1/0,
        synthetic-pool-rs-rolling_shutter/Rolling shutter/1/1,
        synthetic-pool-posenoise-pose_opt/Pose noise/1/2
      } {
        \begin{scope}

        \node[anchor=north west, inner sep=2pt] (label-\i) at (\col*\figurewidth,\row*\figurerowheight) {\tiny\vphantom{g}\label};
        \clip[rounded corners=2pt] (label-\i.south west) -- (label-\i.south east) -- (label-\i.north east) -- (\col*\figurewidth+.99\figurewidth,\figurerowheight*\row) -- (\col*\figurewidth+.99\figurewidth,\figurerowheight*\row-\figureheight) -- (\col*\figurewidth,\figurerowheight*\row-\figureheight) -- cycle;

        \node [inner sep=0,minimum width=\figurewidth,minimum height=\figureheight,fill=red!10!white,anchor=north west] at (\col*\figurewidth, \row*\figurerowheight) {%
            \includegraphics[height=\figureheight]{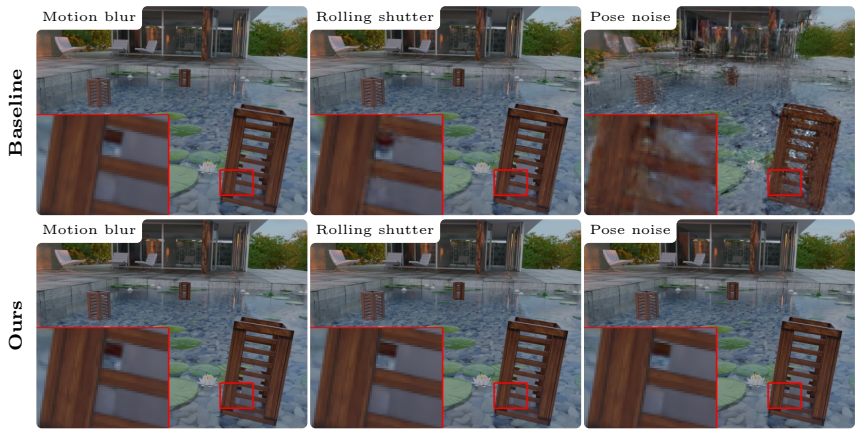}%
        };

        \end{scope}
    }

  \end{tikzpicture}
  \caption{3DGS reconstructions from the synthetic \texttt{pool} scene}
  \label{fig:synthetic-pool}
\end{figure}

\begin{figure}
  \centering\scriptsize

  \begin{tikzpicture}[inner sep=0]

    \setlength{\figurewidth}{.32\textwidth}
    \setlength{\figureheight}{.75\figurewidth}
    \setlength{\figurerowheight}{-1.02\figureheight}

    \node[anchor=center, rotate=90] (label-top) at (-1em,0.5\figurerowheight) {\bf Baseline};
    \node[anchor=center, rotate=90] (label-top) at (-1em,1.5\figurerowheight) {\bf Ours};
    \node[anchor=center, rotate=90] (label-top) at (-1em,2.5\figurerowheight) {\bf Reference};
    \node[anchor=center, rotate=90] (label-top) at (-1em,3.5\figurerowheight) {\bf $l_2$ contrib.};
    \node[anchor=center, rotate=90] (label-top) at (-1em,4.5\figurerowheight) {\bf $\Delta l_2$};

    \foreach \imgtype/\row in {
      baseline/0,
      pose_opt-motion_blur-rolling_shutter-velocity_opt/1,
      gt/2,
      0_l2-ours/3,
      0_l2-diff/4} {
      \foreach \file/\label/\col in {
          colmap-sai-cli-vels-blur-scored-iphone-lego3//0,
          colmap-sai-cli-vels-blur-scored-iphone-pots1//1,
          colmap-sai-cli-vels-blur-scored-s20-sign//2} {
          \begin{scope}

          \node[anchor=north west, inner sep=2pt] (label-\row-\col) at (\col*\figurewidth,\row*\figurerowheight) {\tiny\vphantom{g}\label};

          \clip[rounded corners=2pt] (\col*\figurewidth,\figurerowheight*\row) -- (\col*\figurewidth+.99\figurewidth,\figurerowheight*\row) -- (\col*\figurewidth+.99\figurewidth,\figurerowheight*\row-\figureheight) -- (\col*\figurewidth,\figurerowheight*\row-\figureheight) -- cycle;

          \node [inner sep=0,minimum width=\figurewidth,minimum height=\figureheight,fill=red!10!white,anchor=north west] at (\col*\figurewidth, \row*\figurerowheight) {%
              \includegraphics[height=\figureheight]{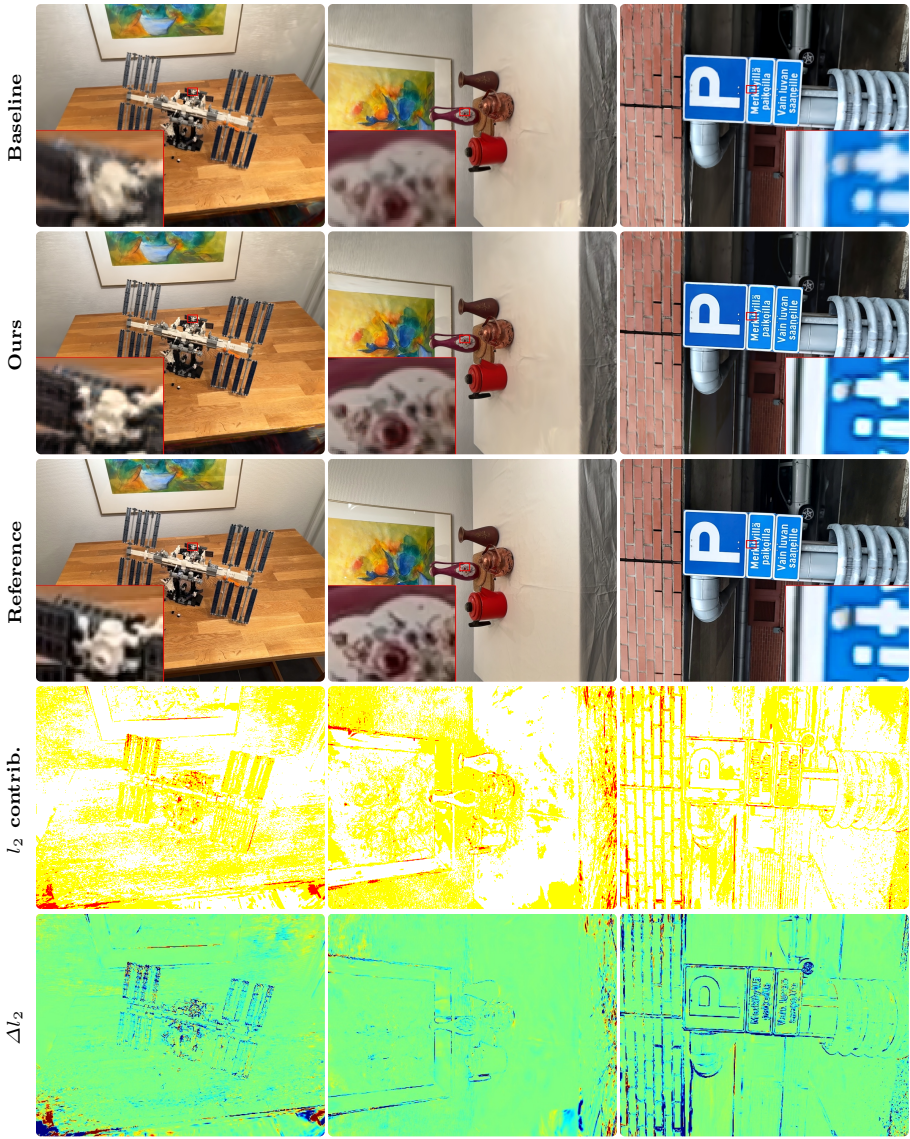}%
          };

          \end{scope}
      }
    }

  \end{tikzpicture}
  \caption{Additional smartphone data reconstructions as in \cref{fig:real} for the scenes \texttt{lego3}, \texttt{pots1} (iPhone) and \texttt{sign} (S20).}
  \label{fig:real3}
\end{figure}

\paragraph{Figures} The synthetic data results for the remaining scenes are visualized in \cref{fig:synthetic-tanabata} and \cref{fig:synthetic-pool}. The reconstructions for rest of the smartphone data sets presented in \cref{tab:real} are shown in \cref{fig:real3}.

\subsubsection{Alternative intrinsics}
In addition to the main sequences, we recorded separate manual camera calibration data for smart phones, using the same fixed focus distance as in the other sequences, and used the Kalibr software package \nopagebreak(\url{https://github.com/ethz-asl/kalibr}) to compute a accurate intrinsic calibration parameters for those devices. We also recorded built-in calibration data reported by the devices.

We then used then trained our method with data where the COLMAP-calibrated intrinsics have been replaced by the manually calibrated intrinsics or the build-in calibration data. The results are presented in \cref{tab:real-additional}. The built-in calibration data reduced the accuracy in all cases, but the manual calibration improved them for some Android data.
We suspect that the reason behind this is that COLMAP also utilizes the camera intrinsic parameters to compensate also for the rolling shutter effect, which results to unoptimal intrinsics solution for a method which can also perform rolling shutter compensation.

We also note that COLMAP-estimated poses do not generally represent the best solution for a given (different) set of camera intrinsics, since slightly inaccurate intrinsics may be compensated by adjusting the poses. Even though this effect can be compensated by our pose optimization, it still appears that for the iPhone data where the rolling shutter effect is negligible, COLMAP-calibrated intrinsics, which are optimized per session, yield better results than manual calibration, which may also have its own inaccuracies.

\paragraph{Blur-based key frame selection}
\cref{tab:real-no-blur-select} shows the results for an experiment where all motion-blur-based scoring for training-evaluation split and key frame selection have been disabled. The results are relatively similar to those in \cref{tab:real}. However, with more blurry evaluation frames, it becomes less clear if the metrics reward sharp reconstructions or accurate simulation of motion blur and rolling shutter effects in the forward model.

\begin{table}[h]
  \centering\scriptsize
  \caption{Smartphone results with alternative  calibration methods. The COLMAP variant data matches 'Ours' in  \cref{tab:real}.\label{tab:real-additional}}
  \setlength{\tabcolsep}{4.5pt}
  \renewcommand*{\arraystretch}{1.2}
  \vspace*{-.1in}
  %------------------------------------------------------------
%------------------------------------------------------------
%------------------------------------------------------------
%
%       AUTO-GENERATED FILE. Edit in generate_tables.py
%
%------------------------------------------------------------
%------------------------------------------------------------
%------------------------------------------------------------
\begin{tabular}{l|ccc|ccc|ccc}
 & \multicolumn{3}{|c}{COLMAP intrinsics} & \multicolumn{3}{|c}{Built-in calibration} & \multicolumn{3}{|c}{Manual calibration}\\
\hline
 & {\tiny \PSNRColName} & {\tiny \SSIMColName} & {\tiny \LPIPSColName} & {\tiny \PSNRColName} & {\tiny \SSIMColName} & {\tiny \LPIPSColName} & {\tiny \PSNRColName} & {\tiny \SSIMColName} & {\tiny \LPIPSColName}\\
\hline
\textsc{iphone-lego1} & \textbf{29.20} & \textbf{.933} & \textbf{.116} & \textit{27.37} & \textit{.903} & \textit{.140} & 27.37 & .902 & .143\\
\textsc{iphone-lego2} & \textbf{27.95} & \textbf{.926} & \textbf{.117} & \textit{27.88} & .917 & \textit{.124} & 27.03 & \textit{.917} & .124\\
\textsc{iphone-lego3} & \textbf{24.50} & \textbf{.824} & \textbf{.203} & 23.58 & \textit{.759} & .250 & \textit{23.83} & .758 & \textit{.238}\\
\textsc{iphone-pots1} & \textbf{29.10} & \textbf{.926} & \textbf{.135} & 28.21 & \textit{.885} & \textit{.171} & \textit{28.25} & .883 & .177\\
\textsc{iphone-pots2} & \textbf{28.00} & \textbf{.882} & \textbf{.180} & 27.53 & \textit{.851} & .198 & \textit{27.68} & .848 & \textit{.197}\\
\textsc{pixel5-lamp} & \textbf{30.46} & \textbf{.833} & \textbf{.196} & 29.50 & .805 & .225 & \textit{30.45} & \textit{.830} & \textit{.199}\\
\textsc{pixel5-plant} & \textbf{27.90} & \textbf{.920} & \textbf{.204} & 27.14 & .902 & .239 & \textit{27.54} & \textit{.919} & \textit{.205}\\
\textsc{pixel5-table} & \textit{31.86} & \textit{.919} & \textit{.196} & 30.60 & .883 & .221 & \textbf{32.26} & \textbf{.919} & \textbf{.194}\\
\textsc{s20-bike} & \textit{28.93} & \textbf{.911} & \textbf{.242} & 28.15 & .887 & .295 & \textbf{29.35} & \textit{.909} & \textit{.246}\\
\textsc{s20-bikerack} & \textbf{29.74} & \textbf{.831} & \textbf{.221} & \textit{27.62} & .814 & .269 & 27.26 & \textit{.827} & \textit{.234}\\
\textsc{s20-sign} & \textit{26.84} & \textbf{.812} & \textit{.206} & 25.31 & .773 & .249 & \textbf{26.84} & \textit{.810} & \textbf{.206}\\
\hline
\textsc{average} & \textbf{28.59} & \textbf{.883} & \textbf{.183} & 27.54 & .853 & .217 & \textit{27.99} & \textit{.866} & \textit{.197}\\
\hline
\end{tabular}

\end{table}

\begin{table}[h]
  \centering\scriptsize
  \caption{Results corresponding to \cref{tab:real} with motion-blur-based key frame selection and training-evaluation split disabled.\label{tab:real-no-blur-select}}
  \setlength{\tabcolsep}{4.5pt}
  \renewcommand*{\arraystretch}{1.2}
  \vspace*{-.1in}
  %------------------------------------------------------------
%------------------------------------------------------------
%------------------------------------------------------------
%
%       AUTO-GENERATED FILE. Edit in generate_tables.py
%
%------------------------------------------------------------
%------------------------------------------------------------
%------------------------------------------------------------
\begin{tabular}{l|ccc|ccc}
 & \multicolumn{3}{|c}{Splatfacto} & \multicolumn{3}{|c}{Ours}\\
\hline
 & {\tiny \PSNRColName} & {\tiny \SSIMColName} & {\tiny \LPIPSColName} & {\tiny \PSNRColName} & {\tiny \SSIMColName} & {\tiny \LPIPSColName}\\
\hline
\textsc{iphone-lego1} & 26.26 & .892 & .202 & \textbf{27.48} & \textbf{.930} & \textbf{.152}\\
\textsc{iphone-lego2} & 27.44 & .914 & .153 & \textbf{27.76} & \textbf{.933} & \textbf{.140}\\
\textsc{iphone-lego3} & 24.16 & .777 & .324 & \textbf{25.63} & \textbf{.853} & \textbf{.198}\\
\textsc{iphone-pots1} & 28.39 & .917 & .202 & \textbf{28.62} & \textbf{.938} & \textbf{.159}\\
\textsc{iphone-pots2} & 28.84 & .878 & .280 & \textbf{29.37} & \textbf{.898} & \textbf{.210}\\
\textsc{pixel5-lamp} & 28.36 & \textbf{.865} & .336 & \textbf{31.57} & .845 & \textbf{.193}\\
\textsc{pixel5-plant} & 26.76 & .933 & .218 & \textbf{28.55} & \textbf{.947} & \textbf{.182}\\
\textsc{pixel5-table} & 28.39 & .916 & .240 & \textbf{31.14} & \textbf{.940} & \textbf{.185}\\
\textsc{s20-bike} & 26.81 & .904 & .288 & \textbf{31.64} & \textbf{.925} & \textbf{.238}\\
\textsc{s20-bikerack} & 26.66 & \textbf{.898} & .258 & \textbf{30.30} & .851 & \textbf{.226}\\
\textsc{s20-sign} & 22.78 & .782 & .290 & \textbf{28.57} & \textbf{.836} & \textbf{.195}\\
\hline
\textsc{average} & 26.80 & .880 & .254 & \textbf{29.15} & \textbf{.900} & \textbf{.189}\\
\hline
\end{tabular}

  \label{tab:without-motion-keyframe-filter}
\end{table}

\begin{figure}[t]
  \centering\scriptsize
  \begin{tikzpicture}[inner sep=0]

    \setlength{\figurewidth}{.24\textwidth}
    \setlength{\figureheight}{.75\figurewidth}
    \setlength{\figurerowheight}{-1.02\figureheight}

    \node[anchor=center, rotate=90] (label-top) at (-1em,0.5\figurerowheight) {\bf Baseline};
    \node[anchor=center, rotate=90] (label-top) at (-1em,1.5\figurerowheight) {\bf Ours};
    \node[anchor=center, rotate=90] (label-top) at (-1em,2.5\figurerowheight) {\bf Reference};
    \node[anchor=center, rotate=90] (label-top) at (-1em,3.5\figurerowheight) {\bf $l_2$ contrib.};
    \node[anchor=center, rotate=90] (label-top) at (-1em,4.5\figurerowheight) {\bf $\Delta l_2$};

    \foreach \imgtype/\row in {
      baseline/0,
      pose_opt-motion_blur-rolling_shutter-velocity_opt/1,
      gt/2,
      0_l2-ours/3,
      0_l2-diff/4} {
      \foreach \file/\label/\col in {
          colmap-sai-cli-vels-blur-scored-iphone-lego2//0,
          colmap-sai-cli-vels-blur-scored-pixel5-plant//1,
          colmap-sai-cli-vels-blur-scored-pixel5-lamp//2,
          colmap-sai-cli-vels-blur-scored-s20-bikerack//3
        } {
          \begin{scope}

          \node[anchor=north west, inner sep=2pt] (label-\row-\col) at (\col*\figurewidth,\row*\figurerowheight) {\label};

          \clip[rounded corners=2pt] (\col*\figurewidth,\figurerowheight*\row) -- (\col*\figurewidth+.99\figurewidth,\figurerowheight*\row) -- (\col*\figurewidth+.99\figurewidth,\figurerowheight*\row-\figureheight) -- (\col*\figurewidth,\figurerowheight*\row-\figureheight) -- cycle;

          \node [inner sep=0,minimum width=\figurewidth,minimum height=\figureheight,fill=red!10!white,anchor=north west] at (\col*\figurewidth, \row*\figurerowheight) {%
              \includegraphics[height=\figureheight]{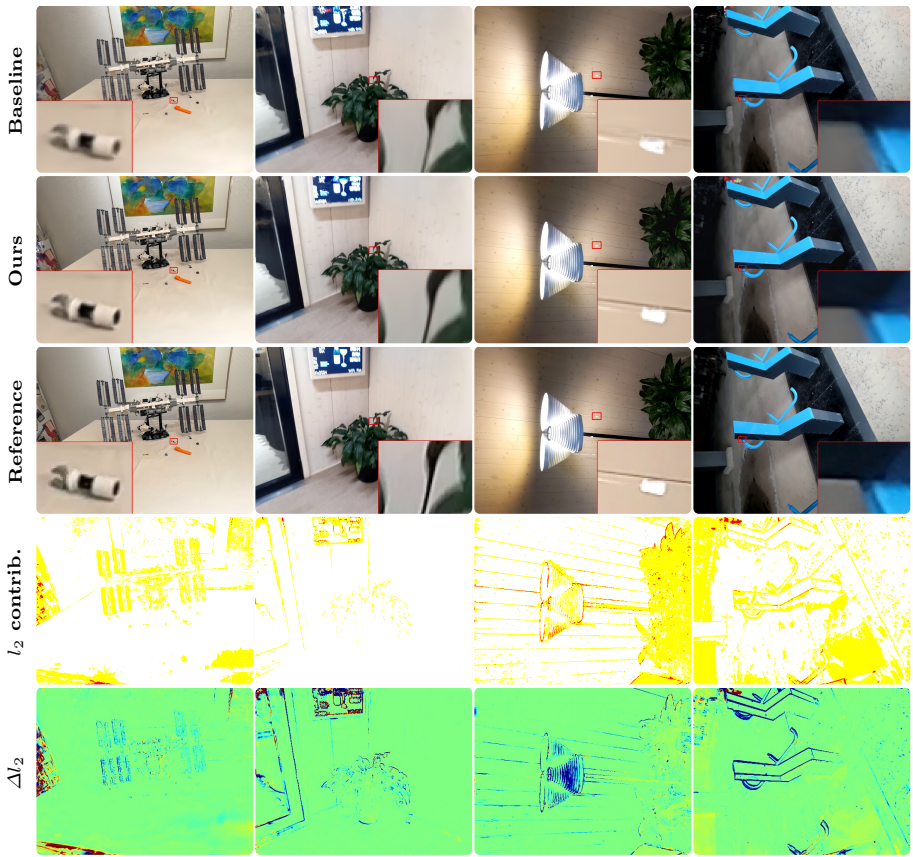}%
          };

          \end{scope}
      }
    }

  \end{tikzpicture}
  \caption{Additional smartphone data reconstructions as in \cref{fig:real} for the scenes \texttt{lego2} (iPhone), \texttt{plant} (Pixel), \texttt{lamp} (Pixel), \texttt{bikerack} (S20).}
  \label{fig:real2}
\end{figure}

\end{document}